\documentclass{article}

\usepackage[margin=3cm]{geometry}
\usepackage{authblk}
\usepackage[pdfencoding=auto, psdextra,hidelinks]{hyperref}
\usepackage{url}
\usepackage[numbers,sort&compress]{natbib}

\input{omri_preamble}
\newcommand{\mytitle}{{Adaptive Symmetrization of the KL Divergence}}

\newif\iftwocolumn
\twocolumnfalse  %

\title{\mytitle{}}

\author[1]{Omri Ben-Dov}
\author[2]{Luiz F.O. Chamon}

\affil[1]{Max Planck Institute for Intelligent Systems, Tübingen AI Center, Tübingen, Germany}
\affil[2]{Department of Applied Mathematics, École Polytechnique de Paris, France}

\date{}

\begin{document}

\maketitle

    \begin{abstract}

The forward Kullback-Leibler (KL) divergence is a ubiquitous objective for fitting a parameterized distribution to samples due to its tractability and equivalence to maximum likelihood estimation (MLE). Its inherent asymmetry, however, may lead to degenerate solutions that generalize poorly. While the symmetric Jeffreys divergence offers a more balanced alternative, its optimization is challenging due to the presence of a reverse KL term. Generative adversarial networks (GANs) bypass this intractability using a min-max formulation at the cost of introducing new instability issues. This work proposes a non-adversarial approach to minimize the Jeffreys divergence. To do so, it uses a proxy model to tractably approximate the reverse KL divergence of the main model. The main and proxy models are jointly fitted to the data using a constrained optimization formulation to obtain a practical algorithm that adapts the models' priorities throughout training. We evaluate our framework on various tasks, including density estimation and simulation-based inference, and demonstrate that this approach is more stable and more accurate than MLE and GANs, particularly in low-data regimes.

\end{abstract}

    \section{Introduction}
\begin{figure*}
    \centering
    \begin{subfigure}{0.3\linewidth}
        \centering              
        \caption{KL minimization}
        \includegraphics[width=\linewidth]{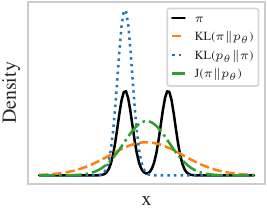}
        \label{fig:mle}

    \end{subfigure}
    \hfill
    \begin{subfigure}{0.3\linewidth}
        \centering      
        \caption{WGAN mode collapse}
        \includegraphics[width=\linewidth]{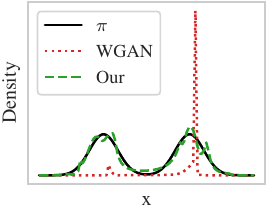}
        \label{fig:teaser_wgan_mode}

    \end{subfigure}
    \hfill
    \begin{subfigure}{0.3\linewidth}
        \centering
        \caption{WGAN instability}
        \includegraphics[width=\linewidth]{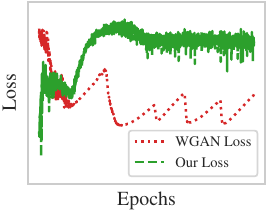}
        \label{fig:teaser_wgan_loss}
        
    \end{subfigure}

    \caption{
        Three advantages of our approach over common learning methods illustrated with a mixture of two Gaussians \pdata{}.
        (a) Fitting a single Gaussian to \pdata{} by minimizing the forward KL leads to a mass-covering behavior (assigning high probability to low-density regions), whereas minimizing the reverse KL leads to a mode-seeking behavior (focusing on a single mode). Our method minimizes the Jeffreys divergence (the sum of forward and reverse KL), balancing these behaviors.
        (b) Training an NF model using the WGAN  objective (a variational form of the Wasserstein metric) leads to mode collapse due to the adversarial training dynamic. In contrast, our framework captures both modes.
        (c) The objective of the WGAN can be unstable, displaying large oscillations, whereas the loss function of our method reaches a stable value.
    }
    \label{fig:teaser}
\end{figure*}

An established approach to learning from observations is to assume they are drawn from a probability distribution, and try to find the distribution that best explains the data~\citep{murphyMachineLearningProbabilistic2012}.
From this learned distribution, we can then solve tasks such as classification, regression, and sampling~\citep{kingmaAutoEncodingVariationalBayes2022,lecunDeepLearning2015,grathwohlYourClassifierSecretly2020}.
Training a neural network (NN) to capture a distribution typically involves minimizing a distance between the ground truth distribution \pdata{}, accessible only through a finite number of samples, and the distribution \pmain{} that the NN induces with parameters \parammain{}.
For instance, parameterized distributions such as normalizing flows (NFs)~\citep{tabakDensityEstimationDual2010,tabakFamilyNonparametricDensity2013,papamakariosNormalizingFlowsProbabilistic2021} and energy-based models (EBMs)~\citep{tehEnergybasedModelsSparse2003,duImplicitGenerationGeneralization2019} are generally trained by minimizing the \emph{forward} Kullback-Leibler~(KL) divergence \(\KL\left(\pdata\parallel \pmain\right)\) between the data distribution \(\pdata\) and the parameterized distribution~\pmain{}.
In practice, these divergences are estimated as an empirical mean over a finite set of samples.
In this case, the forward KL may direct \pmain{} to be a smoother version of \pdata{}, while minimizing the \emph{reverse} KL divergence \(\KL\left(\pmain\parallel \pdata\right)\) may yield a \pmain{} that completely misses some modes of \pdata{} (illustrated in \cref{fig:mle}).
Ordinarily, minimizing a symmetric divergence would alleviate this issue, but it demands knowing the data distribution directly.
Generative adversarial networks (GANs) overcome this obstacle by representing a symmetric divergence as a variational problem involving two models that are trained against each other~\citep{goodfellowGenerativeAdversarialNets2014,nowozinFGANTrainingGenerative2016,arjovskyWassersteinGenerativeAdversarial2017}.
This formulation can provide high-quality results, but is very sensitive to hyperparameters and is prone to mode collapse~(\cref{fig:teaser_wgan_mode}) and instabilities~(\cref{fig:teaser_wgan_loss}).

\subsection{Our contributions}
\paragraph{Non-adversarial Jeffreys divergence minimization.}
In this paper, we seek to minimize the sum of the forward and reverse KL divergences, also known as the (symmetric) \emph{Jeffreys divergence}~\citep{jeffreysTheoryProbability1939}, i.e., \(\jeffreys(\pmain,\pdata) = \KL(\pdata \parallel \pmain) + \KL(\pmain \parallel \pdata)\).
To accomplish this, we use a proxy model \pproxy{} to simulate the data distribution and approximate the reverse KL as \(\KL(\pmain \parallel \pdata) \approx \KL(\pmain \parallel \pproxy)\).
We then use constrained optimization to formulate the joint problem of minimizing the Jeffreys divergence and fitting the proxy model.

\paragraph{Adaptive weighting of objective terms.}
Incorrectly setting the constraint over the proxy model may trivialize the approximation and hinder the optimization of the Jeffreys divergence.
In extreme cases, it may even render the constrained problem infeasible.
We overcome this issue by using \emph{adaptive symmetrization} that dynamically adjusts the priority between optimizing the forward or reverse KL divergence, as well as fitting the proxy model.

\paragraph{Practical algorithm based on (non-convex) duality theory.}
Non-convex constrained problems are generally difficult to optimize.
Therefore, we use duality to solve the constrained problem and define a gradient descent-ascent algorithm to simultaneously train parameterized distributions.
We implement this algorithm with a combination of NFs and EBMs, and show how their simultaneous training can avoid the need for costly EBM sampling by leveraging samples obtained from the NF.
Evaluating on various tasks and datasets, we demonstrate that in the low-data regime, our method maintains training stability and better accuracy than KL minimization and Wasserstein GANs.

    \section{Problem Formulation}

Let \pmain{} be a parameterized distribution with parameters~$\parammain \in \R^k$, such as an NF or an EBM (see \cref{ap:hypotheses}).
We seek \pmain{} that is closest to an unknown data distribution~\pdata{} over~$\R^{\sampledim}$ that is observed only through~$N$ independently and identically distributed~(i.i.d.) samples.
We assume all distributions are absolutely continuous with respect to the Lebesgue measure and use the same notation for a distribution~$p_\parammain$ and its density~\(\pmain\left(x\right)\).

There is a plethora of ways to measure the distance between~\pmain{} and~\pdata{}, from integral probability metrics~(e.g., the Wasserstein distance) to \(f\)-divergences~\citep{renyiMeasuresEntropyInformation1961,aliGeneralClassCoefficients1966,csiszarInformationTypeMeasuresDifference1967}.
While each has its own use cases, the Kullback-Leibler~(KL) divergence~$\KL$~\citep{kullbackInformationSufficiency1951} remains a popular choice as it can be optimized as
\begin{prob}[P-KL] \label{P:kl}
    \minimize_{\parammain\in\R^{k}}\ \KL\left(\pdata\parallel\pmain\right)\triangleq\E_{x\sim\pdata}\left[\log\left(\frac{\pdata\left(x\right)}{\pmain\left(x\right)}\right)\right]=-H\left(\pdata\right)-\E_{x\sim\pdata}\left[\log\pmain\left(x\right)\right]\text{.}
\end{prob}
Note that the entropy~$H(\pdata)=\E_{x\sim \pdata}[-\log \pdata(x)]$, while unknown, is independent of~\parammain{}.
Hence, \eqref{P:kl} is equivalent to a statistical learning problem whose solution can be approximated using empirical risk minimization~(ERM) as in
\begin{prob}[P-MLE]\label{P:mle}
	\minimize_{\parammain\in\R^{k}}\ \nll\left(\pmain\right)\triangleq\ -\dfrac{1}{N}\sum_{i=1}^{N}\log\pmain\left(x_{i}\right).
\end{prob}
Consequently, \eqref{P:kl} is equivalent to maximum likelihood estimation (MLE)~\citep{murphyMachineLearningProbabilistic2012,huberBehaviorMaximumLikelihood1967} or minimum cross-entropy~\citep{shoreAxiomaticDerivationPrinciple1980,rubinsteinCrossEntropyMethod2004}.

Though practical, the KL divergence is not symmetric, i.e., \(\KL\left(\pdata\parallel p\right)\neq \KL\left(p\parallel \pdata\right)\).
Indeed, minimizing the \emph{forward} KL divergence~$\KL\left( \pdata \parallel \pmain\right)$ as in~\eqref{P:kl}, leads to solutions that cover the support of~$\pdata$ at the cost of overlooking sharp modes.
In contrast, the \emph{reverse} KL divergence~$\KL\left( \pmain \parallel \pdata\right)$ promotes a mode-seeking behavior that may ignore regions of \pdata{} with significant mass~\citep{murphyMachineLearningProbabilistic2012} (see \cref{fig:mle} and \cref{ap:kl_behaviors} for details).
One way to overcome this asymmetry is to sum the forward and reverse divergences as in
\begin{prob}[P-JD]
	\label{P:jeffreys}
    \minimize_{\parammain\in\R^{k}}\ \jeffreys(\pmain,\pdata)
    	\triangleq \KL(\pdata \parallel \pmain) + \KL(\pmain \parallel \pdata)
    	\text{.}
\end{prob}
The metric~$\jeffreys$ is referred to as the \emph{Jeffreys divergence} (JD)~\citep{jeffreysTheoryProbability1939,kullbackInformationSufficiency1951}.
Nevertheless, the reverse KL divergence~$\KL(\pmain \parallel \pdata)$ in~\eqref{P:jeffreys} is difficult to estimate and optimize based only on samples, as it requires access to the unknown density \pdata{}.
GANs~\citep{goodfellowGenerativeAdversarialNets2014,arjovskyWassersteinGenerativeAdversarial2017} avoid this challenge by targeting a variational representation of a symmetric divergence, such as the Jensen-Shannon divergence (see \cref{ap:related_work_gan}).
This formulation results in a brittle min-max problem susceptible to instability and to an extreme form of mode-seeking known as \emph{mode collapse}~\citep{arjovskyPrincipledMethodsTraining2017,farniaGANsAlwaysHave2020}.
In contrast, we next tackle the issues facing~\eqref{P:jeffreys} using (i)~a proxy model and (ii)~constrained optimization.

    \section{Minimizing the Jeffreys Divergence}
\label{sec:method}

In this section, we tackle the symmetrized~\eqref{P:jeffreys} by approximating the reverse KL using a proxy model and jointly training the two models using constrained optimization~(\cref{ssec:reversing}).
We then discuss the practical benefits of adapting the level of symmetrization between the weights of the terms in \eqref{P:jeffreys} throughout training.
We enable adaptivity by rewriting \eqref{P:jeffreys} as a feasibility problem with optimizable constraint specifications (\cref{ssec:resilience}).

\subsection{Approximating the reverse KL}
\label{ssec:reversing}

A key roadblock in solving~\eqref{P:jeffreys} is the fact that computing the reverse KL divergence~$\KL(\pmain \parallel \pdata)$ requires integrating over the unknown values of \pdata{} on the support of \pmain{} (see definition in \cref{ap:kl_behaviors}).
Hence, the reverse KL can only be computed after estimating~$\pdata$ from its samples using, e.g., kernel density estimation or \(k\)-nearest neighbors~\citep{silvermanDensityEstimationStatistics2018}.
Though effective for estimation, these methods come with bias-variance trade-offs that are difficult to balance, especially as \pmain{} evolves during training.
Furthermore, this two-step process of first fitting \(\hat{\pdata}\) to then train \pmain{} propagates the variance of the density estimation step to the training of \pmain{}, which can lead to instability and poor performance. 

To overcome these limitations, we incorporate the density estimation step into~\eqref{P:jeffreys}.
Explicitly, consider a proxy distribution model \pproxy{} parameterized by $\paramproxy \in \R^k$ (we assume that \parammain{} and \paramproxy{} have the same dimensionality without loss of generality).
As long as $\pproxy \approx \pdata$, we can use it to train \pmain{} by computing the reverse KL divergence in the objective of \eqref{P:jeffreys} as $\KL(\pmain \parallel \pproxy)$.
Using constrained optimization, we can simultaneously minimize the Jeffreys divergence and ensure that \pproxy{} fits the data. We can also impose additional (optional) constraints, e.g., to regularize the models.
Practically, we replace~\eqref{P:jeffreys} with
\begin{prob}\label{P:proxy}
    \minimize_{\parammain,\paramproxy\in\R^{k}}&&&\KL\left(\pdata\parallel\pmain\right)+\KL\left(\pmain\parallel\pproxy\right)\\
    \subjectto&&&\KL\left(\pdata\parallel\pproxy\right)\leq\epsilon\text{,}\quad h\left(\parammain,\paramproxy\right)\leq 0
    	\text{.}
\end{prob}
Here $\epsilon \geq 0$ determines the accuracy of the proxy model and, thus, of the reverse KL approximation. The function \(h\) defines~(optional) constraints.
Notice that the proxy model \pproxy{} is simultaneously trained to fit~$\pdata$~(as a constraint) as well as used to evaluate the reverse KL of \pmain{} (as the objective).
Also, since~\pproxy{} is included only in forward KL terms, it needs only to estimate density, whereas from~\pmain{} we need both density estimation and sampling, as it is included in the forward and reverse KL forms.
This difference means that in general, \pmain{} and \pproxy{} need not come from the same model family, and it can, in fact, be beneficial to choose different model types~(see~\cref{ssec:ebm_as_proxy}).

\subsection{Adaptive symmetrization}
\label{ssec:resilience}
Depending on the expressiveness of the proxy model \pproxy{}, it may be impossible to choose a small enough value for~$\epsilon$ to ensure the accuracy of the reverse KL.
Even when it is possible, unless we can ensure that \pproxy{} remains close to \pdata{} throughout training, there could be phases in which the reverse KL approximation is inaccurate.
In such cases, we may want to emphasize the forward KL in the objective of~\eqref{P:proxy}, since it is computed directly from data and, hence, more reliable.
In other words, we may want to~(temporarily)~break the symmetry of the Jeffreys divergence during training.
As \pproxy{} starts to better reflect the data, we can go back to balancing the symmetrization.
By dynamically changing the focus between forward and reverse KL divergences, we effectively steer the training toward more mass-covering or more mode-seeking behaviors.

Naturally, it would be impractical to manually adjust these trade-offs during training, especially since they depend in non-trivial ways on the models \pmain{},\pproxy{} and the data.
Attempting to adjust such a schedule manually would lead to time-consuming hyperparameter tuning that would undermine the applicability of this method.
We therefore rely on the resilient optimization framework~\citep{chamonResilientControlCompromising2020,chamonCounterfactualProgrammingOptimal2020,hounieResilientConstrainedLearning2023}.
This approach adjusts constraint specifications such as~$\epsilon$ in~\eqref{P:proxy} depending on how difficult they are to fulfill.
Accordingly, a constraint that is difficult~(easy) to satisfy will have its specification loosened~(tightened). In its simplest form, this behavior can be achieved by writing~\eqref{P:proxy} as a feasibility problem and incorporating the constraint specifications in the objective as
\begin{prob}[P-DYN]
    \label{P:final}
    \minimize_{%
\substack{\parammain,\paramproxy\in\R^{k}\\
\ufor,\urev,\uprx\geq0
}
}&&&\ufor^{2}+\urev^{2}+\uprx^{2}&&
\\ \subjectto&&&\KL\left(\pdata\parallel\pmain\right)\leq\ufor\text{,}&&\KL\left(\pmain\parallel\pproxy\right)\leq\urev\text{,}
\\&&&\KL\left(\pdata\parallel\pproxy\right)\leq\uprx\text{,}&&h\left(\parammain,\paramproxy\right)\leq 0\text{.}
\end{prob}
For simplicity, we keep the constraints of the optional requirements \(h\) fixed.
This formulation is equivalent, under mild conditions, to equalizing the relative sensitivity of the constraints~\citep{chamonResilientControlCompromising2020,chamonCounterfactualProgrammingOptimal2020,hounieResilientConstrainedLearning2023}. 
This sensitivity may lead the forward and reverse KL terms to have differing effective weights.
As a result, \eqref{P:final} does not strictly minimize the Jeffreys divergence, but rather the generalized version
\begin{equation} \label{eq:general_jeffreys}
    \generaljeffreys\left(\pdata\parallel\pmain\right)=\wfw\KL\left(\pdata\parallel\pmain\right)+\left(1-\wfw\right)\KL\left(\pmain\parallel\pdata\right)\text{,}
\end{equation}
where \(\wfw\in\left[0,1\right]\) controls the balance between the terms.

    \section{Algorithm Development}
\label{sec:algorithm}

Solving~\eqref{P:final} is challenging because it still depends on the unknown data distribution~\pdata.
Even if we used the empirical approximation in~\eqref{P:mle}, it would remain a non-convex constrained optimization problem for typical choices of \pmain{} and \pproxy{}, including NFs and EBMs.
As such, \eqref{P:final} is not generally equivalent to the penalty formulation that weighs each term as
\begin{prob}[P-W]\label{P:penalty}
  \minimize_{\parammain,\paramproxy\in\R^{k}}\ %
\wfw\KL\left(\pdata\parallel\pmain\right)+\left(1-\wfw\right)\KL\left(\pmain\parallel\pproxy\right)+\wprx\KL\left(\pdata\parallel\pproxy\right)
\end{prob}
for weights \(\wfw,\wprx > 0\).
These weights define the objective landscape of \eqref{P:penalty}.
In the non-convex setting, different weights can lead to different local minima, or even other global minima, than \eqref{P:final} and \eqref{P:proxy}.
This means that solving one does not necessarily solve the other.
In fact, \eqref{P:penalty} leads to worse results in practice~(see~\cref{fig:weighted}).
To overcome this challenge, we next use results from non-convex duality theory to obtain a practical algorithm.

\subsection{The empirical dual problem}
\label{ssec:duality}

We formulate the empirical version of \eqref{P:final} over the data samples. Explicitly,
\begin{prob}[\(\hat{\text{P}}\)-DYN] \label{P:empirical}
\begin{aligned}P^{\star}\triangleq\minimize_{%
\substack{\parammain,\paramproxy\in\R^{k}\\
\bm{\epsilon},\bm{\delta}\geq0
}
} &  &  & \ufor^{2}-\vfor+\urev^{2}-\vrev+\uprx^{2}-\vprx\\
\subjectto &  &  & \nll\left(\pmain\right)\leq\ufor-\vfor^{2}\text{,} &  & \nll\left(\pproxy\right)\leq\uprx-\vprx^{2}\text{,}\\
 &  &  & \KL\left(\pmain\parallel\pproxy\right)\leq\urev-\vrev^{2}\text{,} &  & h\left(\parammain,\paramproxy\right)\leq 0\text{,}
\end{aligned}
\end{prob}
where we collect~$\ufor,\urev,\uprx$ in the vector~$\bm{\epsilon}$ and $\vfor,\vrev,\vprx$ in the vector~$\bm{\delta}$.
We write~$\bm{\epsilon}, \bm{\delta} \geq 0$ to denote that they belong to the non-negative orthants.
Defining the constraints as a function of both \(\bm{\epsilon}\) and \(\bm{\delta}\) opens the door for negative upper bounds.
These negative bounds, in turn, let us use the NLL of \pmain{} and \pproxy{} rather than the KL divergence, because (i) the constant entropy of \pdata{} does not depend on \parammain{} and \paramproxy{}, and (ii) this constant is absorbed into \(\bm{\epsilon}\) and \(\bm{\delta}\) which are adjusted automatically.
This specific function of \(\bm{\epsilon}\) and \(\bm{\delta}\) leads to a closed-form solution for their optimum, removing the need to compute additional gradients (see \cref{ap:closed_form_epsilon}).

We then define the dual problem of~\eqref{P:empirical} as
\begin{prob}[\(\hat{\text{D}}\)-DYN]\label{D:dualproblem}
	D^{\star}=\maximize_{\lagrangevector\geq0}\ \minimize_{\substack{\parammain,\paramproxy\in\R^{k}\\
\bm{\alleps},\bm{\alldeltas}\geq0
}
}\ %
\Ls\left(\parammain,\paramproxy,\bm{\alleps},\bm{\alldeltas},\lagrangevector\right)
\end{prob}
for the Lagrangian
\begin{equation}
	\label{eq:lagrangian}
  \begin{aligned}\Ls\left(\parammain,\paramproxy,\bm{\epsilon},\bm{\delta},\lagrangevector\right) & =\ufor^{2}-\vfor+\urev^{2}-\vrev+\uprx^{2}-\vprx+\lmbfor\left[-\dfrac{1}{N}\sum_{i=1}^{N}\log\left(\pmain\left(x_{i}\right)\right)-\ufor+\vfor^{2}\right]\\
 & \phantom{=}+\lmbrev\left[\KL\left(\pmain\parallel\pproxy\right)-\urev+\vrev^{2}\right]+\lmbprx\left[-\dfrac{1}{N}\sum_{i=1}^{N}\log\left(\pproxy\left(x_{i}\right)\right)-\uprx+\vprx^{2}\right]\\
 & \phantom{=}+\lambda_{h}\left[h\left(\parammain,\paramproxy\right)\right],
\end{aligned}
\end{equation}

where the dual variables~$\lmbfor,\lmbrev,\lmbprx,\lambda_h$ are collected in the vector~$\lagrangevector\in\R_{+}^{4}$.
While \eqref{D:dualproblem} has a max-min structure, it is not adversarial since both \pmain{} and \pproxy{} minimize the same objective; it is not a zero-sum game.
The maximization over the dual variables \lagrangevector{} simply ensures that \eqref{D:dualproblem} solves the original problem \eqref{P:empirical}.
As seen in \eqref{eq:lagrangian}, this dual objective mirrors the penalty formulation in \eqref{P:penalty}, becoming essentially equivalent when \(\bm{\epsilon}=\bm{\delta} = 0\) and \(\lambda_{h} = 0\).
The crucial distinction is that the dual variables \lagrangevector{} are optimized within the problem, whereas the penalty weights $\bm{w}$ remain fixed.
This distinction allows us to show that~\eqref{D:dualproblem} approximates~\eqref{P:empirical} despite non-convexity.
\begin{theorem}\label{thm:duality_gap}
	Let \(\pmain{}\in\hypotheses_{\parammain}\) and \(\pproxy\in\hypotheses_{\paramproxy}\) such that~$\pmain{},\pproxy{} \geq \eta > 0$ for some \(\eta\), and \(h\) be a convex and Lipschitz continuous function.
	Suppose there exists~$\nu \geq 0$ such that for each~$p \in \bar{\hypotheses}_{\parammain}$, the closed convex hull of~$\hypotheses_{\parammain}$, there exists~$\pmain {\in} \hypotheses_{\parammain}$ such that $\left\|\pmain -  p\right\|_\text{TV} \leq \nu$. Suppose the same holds for~$\hypotheses_{\paramproxy}$. Then, there exists a finite constant~$B$ such that
	\begin{equation}\label{eq:duality_gap}
	    0 \leq P^\star - D^\star \leq B \nu
	    	\text{.}
	\end{equation}
\end{theorem}
\begin{proof}
	The result follows from~\citet[Prop. III.3, Lemma III.1]{chamonConstrainedLearningNonConvex2023} by noting that (i)~for~$\pmain , \pproxy \geq \eta > 0$, the losses in~\eqref{P:empirical} are convex and Lipschitz continuous~(Assumption 1 in Prop. III.3), and (ii)~$\hypotheses_{\parammain}$ and~$\hypotheses_{\paramproxy}$ satisfy Assumption 3 by hypothesis, and (iii) Assumption 4, Slater's condition, is satisfied trivially since~$\bm{\epsilon}$ is an optimization variable.  Note that Assumption~2 is not needed for either result.
\end{proof}

\begin{wrapfigure}{r}{0.45\textwidth} %
	\vspace{-20pt}
  \begin{minipage}{\linewidth} %
	\begin{algorithm}[H]
  \caption{Gradient descent-ascent}
  \label{alg:pseudocode}
  \begin{algorithmic}
	\REQUIRE Learning rates \(\lr_{\parammain},\lr_{\paramproxy},\lr_{\eps},\lr_{\delta},\lr_{\lambda}\)
    \REPEAT
    \STATE \(\parammain\leftarrow\parammain-\lr_{\parammain}\nabla_{\parammain}\Ls \)
    \STATE \(\paramproxy\leftarrow\paramproxy-\lr_{\paramproxy}\nabla_{\paramproxy}\Ls \)
	\STATE \(\alleps\leftarrow\max\left\{ \bm{0},\alleps-\lr_{\eps}\nabla_{\alleps}\Ls\right\} \)
	\STATE \(\alldeltas\leftarrow\max\left\{ \bm{0},\alldeltas-\lr_{\delta}\nabla_{\alldeltas}\Ls\right\} \)
	\STATE \(\lagrangevector\leftarrow\max\left\{ \bm{0},\lagrangevector+\lr_{\lambda}\nabla_{\lagrangevector}\Ls\right\} \)
	\UNTIL{convergence}
	\end{algorithmic}
	\end{algorithm}
  \end{minipage}
  \vspace{-10pt}
  \end{wrapfigure}
\Cref{thm:duality_gap} enables us to (approximately)~solve the constrained problem~\eqref{P:empirical} by means of its unconstrained dual problem~\eqref{D:dualproblem}.
The latter can be tackled using gradient descent-ascent~(GDA) techniques common in primal-dual methods~\citep{boydConvexOptimization2004,chamonProbablyApproximatelyCorrect2020,chamonConstrainedLearningNonConvex2023,elenterNearoptimalSolutionsConstrained2024}.
They involve taking~(projected)~gradient descent steps for the primal optimization variables~$\parammain,\paramproxy,\alleps,\alldeltas$ and~(projected)~gradient ascent steps for the dual variables~$\lagrangevector$~(see~\Cref{alg:pseudocode}). \Cref{thm:duality_gap} also shows that the approximation error can be made arbitrarily small by using sufficiently rich models for \pmain{} and \pproxy{}.

    \section{Implementation and Experimental Results}
\label{sec:experiments}

\begin{figure*}
    \centering

    \begin{subfigure}{0.49\linewidth}
        \centering

        \caption{UCI Breast Cancer}
        \includegraphics[width=\linewidth]{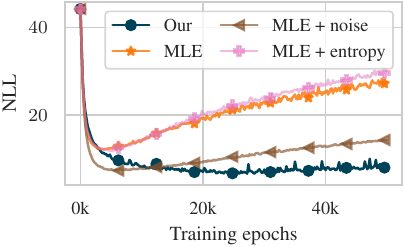}
        \label{fig:cancer}

    \end{subfigure}
    \hfill
    \begin{subfigure}{0.49\linewidth}
        \centering
        
        \caption{UCI Heart Disease}
        \includegraphics[width=\linewidth]{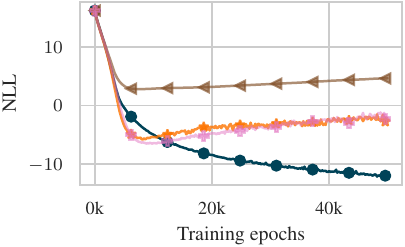}
        \label{fig:heart}
        
    \end{subfigure}

    \caption{
        Our framework outperforms MLE approaches on UCI datasets. Our method achieves lower and more stable NLL, while MLE approaches usually diverge due to overfitting.
        Adding random noise to the samples for MLE can mitigate overfitting (left), but is very sensitive to the amount of noise which can lead to worse accuracy (right).
    }
    \label{fig:uci}
\end{figure*}

\subsection{Comparison with MLE}
\label{ssec:results_uci}
We compare our method with MLE baselines on three UCI datasets.
We implement our framework with NFs (\cref{ap:nf_details}) for \pmain{} and \pproxy{} because NFs are quick and easy to train, unlike the more compute-intensive EBMs (\cref{ap:ebm_details}).
As we lack access to the ground truth distribution, we cannot compute the actual Jeffreys divergence and instead show the NLL of the different methods on the test sets throughout training (\cref{fig:uci}).
\cref{ap:mle_noise} includes additional results, details about the datasets and pre-processing, and technical training configurations.

\paragraph{Noise perturbation.}
We first compare solving \eqref{D:dualproblem} (Ours) with solving \eqref{P:mle} (MLE).
\cref{fig:uci} shows that after the first few iterations, MLE diverges from the true distribution due to overfitting (\cref{ap:kl_behaviors}), while our method remains stable at a lower NLL.
One approach to mitigate overfitting (at the cost of smoothing the distribution) is to add random noise to the data samples~(MLE + noise)~\citep{bishopTrainingNoiseEquivalent1995}.
While carefully designed noise can improve the performance of MLE in some cases (as in \cref{fig:cancer}), the same noise may lead to higher NLL in other cases (as in \cref{fig:heart}).
We present further evidence of noise sensitivity and overfitting in \cref{ap:mle_noise}.

\paragraph{Maximum entropy.}
The reverse KL term includes the negative entropy of \pmain{}, which may help avoid overfitting.
As an ablation, we also train MLE with regularization to maximize the entropy of \pmain{}~(MLE + entropy).
As \cref{fig:uci} shows, adding this regularization term does not prevent overfitting and behaves similarly to the standard MLE.
This serves as evidence that the advantages of our framework are not solely due to the additional entropy term in the reverse KL.

\begin{figure*}
    \centering

        \begin{subfigure}{0.49\linewidth}
        \centering      
        \caption{Jeffreys divergence}
        \label{fig:weighted_psi_jeffreys}

        \includegraphics[width=\linewidth]{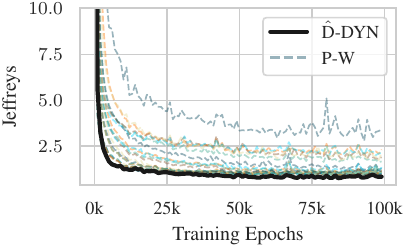}

    \end{subfigure}
    \hfill
    \begin{subfigure}{0.49\linewidth}
        \centering
        \caption{Estimated \partition{}}
        \label{fig:weighted_zeta}

        \includegraphics[width=\linewidth]{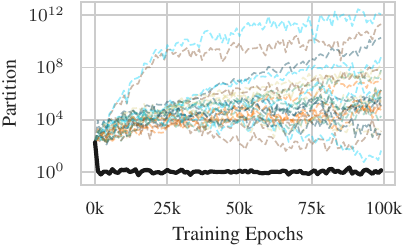}

    \end{subfigure}

    \caption{
        Solving the dual problem \eqref{D:dualproblem} (solid black line) achieves better results than 25 weight configurations of the weighted problem \eqref{P:penalty} (dashed colored lines) over a synthetic Gaussian mixture dataset.
    (a) Solving the dual problem converges to better Jeffreys divergence than any weight configuration.
    (b) With further constraints, \eqref{D:dualproblem} maintains the partition function of the EBM \pproxy{} close to unity, while the weighted problem may reach large unstable values.
    }
    \label{fig:weighted}
\end{figure*}

\subsection{Efficient EBM training as proxy}
\label{ssec:ebm_as_proxy}
Since NFs have to be bijective functions, they are restricted to a limited set of architectures.
In contrast, EBMs have no such restrictions, but they represent unnormalized distributions (see \cref{ap:ebm_details}).
Since EBMs are unnormalized, evaluating and optimizing their log-likelihoods require computing their partition function \(\zeta\).
The partition function is usually evaluated by obtaining (a potentially large number of)~samples from the EBM's distribution, which calls for time-consuming Markov Chain Monte Carlo (MCMC) methods.
We propose to use an NF for \pmain{} and an EBM for the proxy \pproxy{} to exploit their complementary nature and overcome their limitations.
This symbiotic relationship between NFs and EBMs has been observed and exploited in prior works (see \cref{ap:related_work_joint}).
Since it is straightforward to both obtain samples and their densities from \pmain{} as an NF, the partition function of the EBM \pproxy{} can therefore be estimated using importance sampling as in
\begin{equation}\label{eq:partition_is}
	\partition=\int e^{\funcproxy\left(y\right)}\text{d}y=\int e^{\funcproxy\left(y\right)}\frac{\pmain\left(y\right)}{\pmain\left(y\right)}\text{d}y=\E_{y\sim\pmain}\left[e^{\funcproxy\left(y\right)-\log\pmain\left(y\right)}\right]\text{.}
\end{equation}
This replaces the slow EBM sampling by the faster NF, though at the cost of a potentially higher estimator variance. Yet, notice that~$\urev$ in~\eqref{P:empirical} has the effect of keeping \pmain{} and \pproxy{} close to each other, reducing the variance of the importance sampling estimate in~\eqref{eq:partition_is}~\citep{kloekBayesianEstimatesEquation1978,chatterjeeSAMPLESIZEREQUIRED2018,sanz-alonsoImportanceSamplingNecessary2018}.
We can further improve the numerical stability by ensuring that~$\partition$ remains close to one, i.e., that \pproxy{} is almost normalized.
This is possible by setting \(h\left(\parammain,\paramproxy\right)=\left[\partition-1-\upartition,\ 1-\partition\right]^{\top}\) in~\eqref{P:empirical}, showcasing the flexibility of our approach.
These observations yield a practical algorithm to simultaneously train an NF and an EBM~(see~\cref{ap:algorithm} for a detailed description of the algorithm).
In the rest of the paper we use this NF-EBM combination as the main and proxy models.

\paragraph{Comparison with fixed weights.} 
We compare our method to the penalty approach \eqref{P:penalty} with fixed weights.
To that end, we train both formulations on a synthetic 2D dataset of 800 points drawn from a GMM with 40 components~(see \cref{ap:gmm40} for the technical details of this experiment). 
We estimate the Jeffreys divergence of \pmain{} and the partition function of the EBM \pproxy{} over a test dataset of 100k samples.
We compare 25 different weight configurations for \eqref{P:penalty} against a single instance of \eqref{D:dualproblem} in \cref{fig:weighted_psi_jeffreys}.
The adaptive weights of the dual problem attain lower Jeffreys divergence than any of the weight combinations we tried.
Additionally, \cref{fig:weighted_zeta} illustrates how the constraints of \partition{} in \cref{D:dualproblem} make it more numerically stable than \eqref{P:penalty}.
This constraint effectively controls the value of \partition{}, which in contrast can become large when solving \eqref{P:penalty}.
Though this instability can be addressed by other means, such as regularization or projected gradient methods, they often require additional hyperparameters to tune.
\begin{figure}
    \centering

    \begin{subfigure}{0.49\linewidth}
        \centering      
        \caption{\(\lr_{\parammain}=10^{-3}\)}
        \label{fig:wgan_nll3}

        \includegraphics[width=\linewidth]{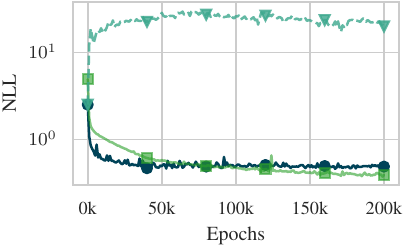}

    \end{subfigure}
    \hfill
    \begin{subfigure}{0.49\linewidth}
        \centering
        \caption{\(\lr_{\parammain}=10^{-5}\)}
        \label{fig:wgan_nll5}

        \includegraphics[width=\linewidth]{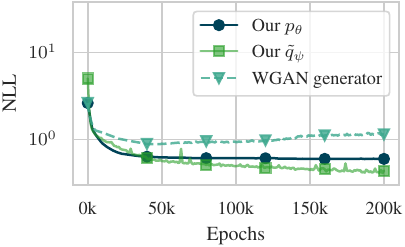}

    \end{subfigure}

    \caption{
    Solving \eqref{D:dualproblem} results in better NLL than WGAN on a synthetic 2D GMM dataset, and is more robust to changing learning rates.
    (a) With a high learning rate, our approach converges to a low NLL, while WGAN diverges. 
    (b) At lower learning rates, WGAN can converge to low NLL, but our approach still achieves better NLL.
    }
    \label{fig:wgan}
\end{figure}

\paragraph{Comparison with WGANs.}
We also compare our method to a Wasserstein GAN with gradient penalty (WGAN-GP)~\citep{gulrajaniImprovedTrainingWasserstein2017}, which targets the symmetric Wasserstein distance.
This comparison uses the same architecture and learning rate for \pmain{} and the WGAN's generator and the same architecture and LR for \pproxy{} and the WGAN's discriminator.
This is a fair comparison since the discriminator in a GAN is in fact an implicit EBM~\citep{cheYourGANSecretly2020,ben-dovAdversarialLikelihoodEstimation2024}.
In \cref{fig:wgan} we present the NLL throughout training of our \pmain{}, \pproxy{} and the WGAN generator.
Since we approximate the partition function of \pproxy{}, their exact Jeffreys divergence is unknown and the values we report are estimates.
To make this fact explicit, we denote the quasi-normalized density of \pproxy{} as \(\tilde{q}_{\paramproxy}\).
As expected, WGAN is stable only at low learning rates (\cref{fig:wgan_nll5}) due to the opposing objectives of the generator and the discriminator.
Conversely, our collaborative approach remains stable over a wider range of LRs and reaches lower NLL values.
In \cref{ap:gmm40} we include additional comparisons of more learning rates, and the evolution of the dual variables over time.
We also compare the training time of the methods in \cref{ap:compute_time} and show that the overhead of our approach with an EBM proxy is comparable to that of WGAN-GP.

\begin{figure*}
    \centering

    \begin{subfigure}{0.45\linewidth}
        \centering
        \caption{Concentric rings NLL}
        \label{fig:vals_concentric}
        \includegraphics[width=\linewidth]{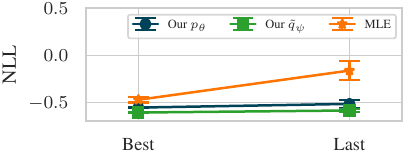}

    \end{subfigure}
    \hfill
    \begin{subfigure}{0.45\linewidth}
        \centering      
        \caption{Concentric rings density}
        \label{fig:density_concentric}
        \includegraphics[width=\linewidth]{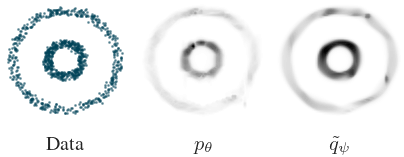}

    \end{subfigure}

    \begin{subfigure}{0.45\linewidth}
        \centering
        \caption{Gaussian mixture grid NLL}
        \label{fig:vals_grid}
        \includegraphics[width=\linewidth]{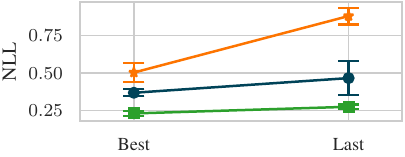}

    \end{subfigure}
    \hfill
    \begin{subfigure}{0.45\linewidth}
        \centering      
        \caption{Gaussian mixture grid density}
        \label{fig:density_grid}
        \includegraphics[width=\linewidth]{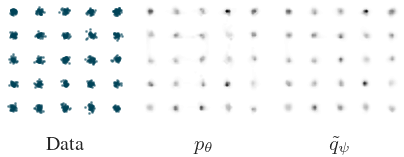}

    \end{subfigure}

    \caption{
    Our framework is able to accurately learn the density of various 2D datasets.
    The left column (a,c) presents the lowest NLL each model achieved during training (``Best'') and the NLL at the end of training (``Last''). The error bars represent the standard deviation over 5 runs with different random seeds, demonstrating that our method outperforms NF in terms of values and consistency.
    The right column (b,d) qualitatively displays the finite dataset (Data), the learned density \pmain{}, and the quasi-normalized density \(\tilde{q}_{\paramproxy}\).
    The qualitative density maps show that both models perfectly capture the shape and all modes.
    }
    \label{fig:main_density}
\end{figure*}

\subsection{Density estimation}
\label{ssec:density}
We test the density estimation capabilities of our method with an EBM proxy on five common 2D datasets: Gaussian mixture ring, Gaussian mixture grid, two moons, spiral, and concentric circles.
\cref{fig:main_density} presents two of these datasets, while \cref{ap:shapes} presents the rest and details the training configurations.
Across all datasets, our method learns a density with lower and more consistent NLL than a pure NF (\cref{fig:vals_concentric,fig:vals_grid}).
Moreover, the qualitative density maps in \cref{fig:density_concentric,fig:density_grid} show that both models perfectly capture the shape and all modes.
In \cref{ap:img_sampling} we also include experiments on sampling from the CelebA dataset.

\begin{figure*}
    \centering

    \begin{subfigure}{0.49\linewidth}
        \centering      
        \caption{Gaussian mixture}

        \includegraphics[width=\linewidth]{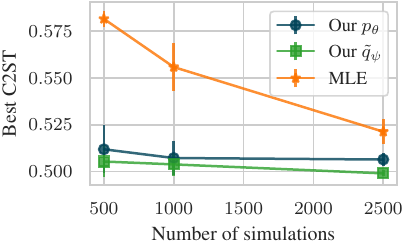}
    \end{subfigure}
    \hfill
    \begin{subfigure}{0.49\linewidth}
        \centering      
        \caption{Two moons}

        \includegraphics[width=\linewidth]{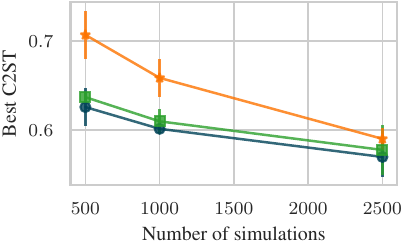}
    \end{subfigure}

    \caption{
    Our method requires less simulations than NF to achieve C2ST scores closer to 0.5 for common SBI benchmarks.
    For each method, over 5 different seeds, we compute the best C2ST over the run and report the mean and standard deviation as the error bars.
    }
    \label{fig:sbi_c2st}
\end{figure*}

\subsection{Simulation-based inference}
\label{ssec:sbi}
Here, we use our dual formulation to learn conditional distributions for simulation-based inference (SBI).
Determining the parameters \sbiparam{} of a mechanism given observations of its outcome \sbioutcome{} is a classical task in scientific applications.
However, even with perfect knowledge of the mechanism and the ability to predict \(\pdata\left(\sbioutcome|\sbiparam\right)\), the inverse, i.e., inferring the parameters given the outcome \(\pdata\left(\sbiparam|\sbioutcome\right)\), can be challenging.
For example, despite having a perfect model to predict gravitational waves from a known source, it is harder to infer the sources given a wave~\citep{daxRealtimeGravitationalWave2021,wildbergerAdaptingNoiseDistribution2023}.
SBI provides solutions to amortized inference by training a model on simulated instances~\citep{cranmerFrontierSimulationbasedInference2020}.
A simple method to perform this inference is neural posterior estimation (NPE)~\citep{papamakariosFastEfreeInference2016,lueckmannFlexibleStatisticalInference2017,radevBayesFlowLearningComplex2022}, where a prior distribution \(p_{\sbiparam}\) over the parameters is selected, \(N\) instances of \sbiparam{} are sampled, and \(N\) outcomes \(\pdata\left(\sbioutcome|\sbiparam\right)\) are simulated.
The conditional inference network then trains on the pairs \((\sbiparam,\sbioutcome)\) to learn the posterior \(\pdata\left(\sbiparam|\sbioutcome\right)\).
Prior works suggest that maximizing the entropy of the learned distribution improves the chances of recovering the underlying distribution~\citep{vetterSourcererSamplebasedMaximum2024}.
Note that this regularization is not needed when targeting the Jeffreys divergence, since the Lagrangian (\cref{eq:lagrangian}) already accounts for the entropy of \pmain{} naturally.

In our experiments, we use a conditional version of \eqref{P:empirical}, described in \cref{ap:conditional} with an EBM proxy.
We trained our method on the common SBI benchmarks two moons~\citep{greenbergAutomaticPosteriorTransformation2019} and Gaussian mixture~\citep{sissonSequentialMonteCarlo2007}.
As is typical in the SBI literature~\citep{rameshGATSBIGenerativeAdversarial2022,vetterSourcererSamplebasedMaximum2024}, we measure the effect of the size of the dataset (number of simulations) on the quality of the learned posterior (see \cref{ap:sbi} for the full technical details).
For evaluation, we used the classifier-two-sample test (C2ST).
The C2ST score is defined as the accuracy of a classifier trained to distinguish between samples from the ground truth posterior and samples from the learned posterior.
The optimal value is 0.5, indicating that the generated posterior is indistinguishable from the ground truth posteriors.
To train the classifier, we created a dataset by sampling 5000 points from both the ground truth posterior and our models, conditioning on the same predetermined \(\sbioutcome_{0}\), and labeling them accordingly.
We ran the experiments 5 times with different seeds. 
\cref{fig:sbi_c2st} shows that our method generates higher quality posterior samples than an NF trained on the same number of samples.

    \section{Conclusion}
Common methods to learn distributions from finite samples include MLE, which may result in mass-covering solutions, and GANs, whose unstable min-max objective may not converge under certain hyperparameters.
In this paper, we minimize the symmetric Jeffreys divergence by having two models collaborate rather than being adversarial, leading to a balance between mass-covering and mode-seeking behaviors, and robustness to hyperparameters.
We do so by jointly training a proxy model to estimate the reverse KL divergence.
Then, by formulating the learning problem as an adaptive constrained optimization problem, we devise a practical algorithm that dynamically shifts focus between the different terms.
We demonstrate on various tasks that our method is stable and requires fewer data points to arrive at better accuracy than MLE and GANs.
Our experiments are limited to low-dimensional benchmarks, though this restriction results from the computational complexity of training NFs and EBMs in higher dimensions.
This problem may be mitigating by using specialized methods~\cite{cateriniRectangularFlowsManifold2021,ben-dovAdversarialLikelihoodEstimation2024}.
Moreover, we demonstrate how our approach can jointly train an NF with an EBM, to efficiently train the EBM without taxing MCMC sampling.

We believe our approach hints that there is still much to explore in the domain of statistical divergences and minimization methods.
Our method can also be extended to other types of divergences, such as the Fisher divergence, which compares the score functions of the distributions~\cite{fisherTheoryStatisticalEstimation1925,hyvarinenEstimationNonNormalizedStatistical2005}.

{}

\subsubsection*{Acknowledgments}
OB is supported by the International Max Planck Research School for Intelligent Systems (IMPRS-IS) with the help of Samira Samadi.
OB conducted this work at the École polytechnique, Paris, and was funded by the Machine Learning Cluster of Excellence, EXC number 2064/1 - Project number 390727645.

\bibliographystyle{unsrtnat}
\bibliography{references}

\appendix
\crefalias{section}{appendix}
\crefalias{subsection}{appendix}
\crefalias{subsubsection}{appendix}
\section{Related Work}

\subsection{Hypothesis classes} \label{ap:hypotheses}
There are as many ways to define~$\hypotheses$ as there are ways of defining a probability distribution.
A traditional choice from variational Bayesian inference considers a set of parameterized distributions, e.g., a Gaussian mixture model~(GMM)~\citep{mclachlanFiniteMixtureModels2000}.
A more flexible approach used by \emph{normalizing flows}~(NFs) consists of letting~$p_\parammain \in \hypotheses$ be the pushforward of a tractable distribution~$p_0$ by an invertible function~$g_\parammain: \R^{\sampledim} \to \R^{\sampledim}$, namely,
\begin{equation}\label{eq:nf}
    p_\parammain(x)
    	= p_0\big( g_\parammain^{-1}(x) \big) \bigl\vert \det D g_\parammain^{-1}(x) \bigr\vert
\end{equation}
where~$D g_\parammain^{-1}$ denotes the Jacobian of the inverse of~$g_\parammain$.
While~$g_\parammain$ is often a neural network~(NN), its architecture has to be carefully designed to ensure that its inverse and Jacobian are easy to compute~\citep{dinhDensityEstimationUsing2017}.
This restriction reduces the flexibility of \hypotheses{}, but has the advantage of allowing NFs to efficiently sample from the distribution~$p_\parammain$ \emph{as well as} evaluate its density.

A less restricted class is that of \emph{energy-based models}~(EBMs) that are based on unnormalized representations of distributions. Explicitly, EBMs are parameterized by an arbitrary function~$f_\parammain: \R^{\sampledim} \to \R$ that implies the density
\begin{equation}\label{eq:ebm}
	p_\parammain(x) = \frac{ e^{f_\parammain(x)} }{ \zeta_\parammain }
		\text{,} \quad \text{ with } \zeta_\parammain = \int e^{f_\parammain(y)} \text{d}y
		\text{.}
\end{equation}
Since there are no restrictions on~$f_\parammain$, EBMs offer more powerful distribution models, though it is typically intractable to evaluate their density~$p_\parammain$ since estimating the partition function~$\zeta_\parammain$ in~\eqref{eq:ebm} is difficult even in moderate dimensions~\citep{kloecknerApproximationFinitelySupported2012}.
Nevertheless, it is possible to obtain samples directly from~$f_\parammain$ using Markov Chain Monte Carlo~(MCMC) methods such as the Langevin Monte Carlo~(LMC) algorithm~\citep{parisiCorrelationFunctionsComputer1981,grenanderRepresentationsKnowledgeComplex1994}.
Further details about NF and EBM are provided in \cref{ap:nf_details,ap:ebm_details}.

\subsection{Normalizing flows}
\label{ap:nf_details}
Unlike EBMs, which directly define the unnormalized density, NFs define an invertible mapping \(g_{\parammain}:\R^{\sampledim}\to\R^{\sampledim}\), parameterized by \parammain{}, from a simple base distribution \basedist{} to the data distribution~\citep{tabakDensityEstimationDual2010,tabakFamilyNonparametricDensity2013,danilorezendeVariationalInferenceNormalizing2015,papamakariosNormalizingFlowsProbabilistic2021}.
Training NF involves minimizing the NLL from \cref{P:kl} through the change-of-variables law of probabilities
\begin{equation}
    \log\pmain\left(x\right)=\log\basedist\left(g_{\parammain}^{-1}\left(x\right)\right)+\log\left|\det\frac{\text{d}g_{\parammain}^{-1}\left(x\right)}{\text{d}x}\right|,
\end{equation}
where \(\frac{\text{d}g_{\parammain}^{-1}\left(x\right)}{\text{d}x}\) is the Jacobian of \(g_{\parammain}^{-1}\) at \(x\).
While computing the Jacobian of a multivariate function is computationally expensive, practical NF relies on architectures with tractable Jacobians~\citep{dinhDensityEstimationUsing2017,kingmaGlowGenerativeFlow2018,durkanNeuralSplineFlows2019}.
These Jacobians simplify the training of NF, and sampling is quick compared to EBMs.
Yet, since the forward KL is minimized from a finite dataset, NFs tend to overfit to the empirical distribution rather than the underlying distribution.
This problem is usually addressed by adding random noise to the data~\citep{uriaRNADERealvaluedNeural2013,hoFlowImprovingFlowbased2019}.
Sampling from NF is quicker than sampling from an EBM as it requires only sampling from the base distribution and applying the invertible mapping.

\subsection{Energy-based models}
\label{ap:ebm_details}
Any function \(\funcproxy:\R^{\sampledim}\to\R\) has a corresponding probability distribution
\begin{equation}
    \label{eq:ebm2}
    \begin{aligned} & \pproxy\left(x\right)=\frac{e^{\funcproxy\left(x\right)}}{\partition}, \text{ with} & \partition=\int e^{\funcproxy\left(y\right)}\text{d}y.\end{aligned}
\end{equation}
The exponent keeps the output non-negative and the partition function \partition{} ensures that the distribution integrates to one.
Essentially, \cref{eq:ebm2} is the definition of a Boltzmann distribution with an energy function \(-\funcproxy\), consequently naming this class of functions as energy-based models~\citep{tehEnergybasedModelsSparse2003,jianwenxieTheoryGenerativeConvNet2016,duImplicitGenerationGeneralization2019}.

The gradient of the NLL objective in \cref{P:kl} with \pproxy{} from \cref{eq:ebm2} is
\begin{equation}
    \label{eq:ebm_grad}
    \begin{aligned}\nabla_{\paramproxy}\E_{x\sim\pdata}\left[-\log\pproxy\left(x\right)\right] & =\E_{x\sim\pdata}\left[-\nabla_{\paramproxy}\funcproxy\left(x\right)+\nabla_{\paramproxy}\log\partition\right]\\
 & =\E_{x\sim\pdata}\left[-\nabla_{\paramproxy}\funcproxy\left(x\right)\right]+\nabla_{\paramproxy}\log\partition\\
 & =\E_{x\sim\pdata}\left[-\nabla_{\paramproxy}\funcproxy\left(x\right)\right]+\frac{1}{\partition}\nabla_{\paramproxy}\int e^{\funcproxy\left(y\right)}\text{d}y\\
 & =\E_{x\sim\pdata}\left[-\nabla_{\paramproxy}\funcproxy\left(x\right)\right]+\int\frac{1}{\partition}e^{\funcproxy\left(y\right)}\nabla_{\paramproxy}\funcproxy\left(y\right)\text{d}y\\
 & =\E_{x\sim\pdata}\left[-\nabla_{\paramproxy}\funcproxy\left(x\right)\right]+\int\pproxy\left(y\right)\nabla_{\paramproxy}\funcproxy\left(y\right)\text{d}y\\
 & =\E_{x\sim\pdata}\left[-\nabla_{\paramproxy}\funcproxy\left(x\right)\right]+\E_{y\sim\pproxy}\left[\nabla_{\paramproxy}\funcproxy\left(y\right)\right]
\end{aligned}
\end{equation}
The second term of the gradient requires sampling from the EBM itself~\citep{duImplicitGenerationGeneralization2019,songHowTrainYour2021}.
Sampling from \pproxy{} is not straightforward and typically involves running a Markov Chain Monte Carlo (MCMC) method, such as Langevin dynamics~\citep{parisiCorrelationFunctionsComputer1981,grenanderRepresentationsKnowledgeComplex1994}.
Starting with \(x_{0}\) from a simple prior distribution, the \(k\)-th Langevin step is 
\begin{equation}
    \label{eq:mcmcsampling}
    x_{k}=x_{k-1}+s\nabla_{x}\log\pproxy\left(x\right)+\sqrt{2s}\vz_{k-1},
\end{equation}
with \(s\) being the step-size and \(\vz\sim\normaldist\).
EBMs are a powerful class of functions due to their unrestricted architecture, but the Langevin sampling makes their training slow and requires more hyperparameter tuning.

\subsection{Mass-covering and mode-seeking behaviors of minimizing KL divergences}
\label{ap:kl_behaviors}
Minimizing the forward KL, as in \eqref{P:kl}, does not guarantee a mass-covering solution, yet the objective encourages this behavior due to the term \(\E_{x\sim\pdata}\left[\log\left(\frac{\pdata\left(x\right)}{\pmain\left(x\right)}\right)\right]\).
As the expectation is over~\pdata{}, the numerator \(\pdata\left(x\right)>0\) is always positive while the denominator \(\pmain\left(x\right)\) can be very small or even zero, leading to large ratios.
In certain cases, when the training data is a finite set, \pmain{} may focus only on the samples with minuscule ratios (high likelihood), ending up with a support that includes only training samples, which is a form of overfitting.
Otherwise, to avoid the high (or even infinite) penalty of missing a sample, \pmain{} may assign significant probability to the regions between samples that may have zero probability under \pdata{}.
As a result, the density of \pmain{} can be a smoothed version of \pdata{}, sacrificing precision to ensure total coverage of the training set (\cref{fig:mle}).

Similarly, minimizing the reverse KL, 
\begin{equation}
    \KL\left(\pmain\parallel\pdata\right)=\E_{x\sim\pmain}\left[\log\left(\frac{\pmain\left(x\right)}{\pdata\left(x\right)}\right)\right]
\end{equation}
does not ensure a mode-seeking solution, but incentivizes it.
In the case where the support of \pmain{} includes regions outside of the support of \pdata{}, the ratio \(\frac{\pmain\left(x\right)}{\pdata\left(x\right)}\) can be very large (or even infinite), leading to a large penalty.
To avoid this penalty, \pmain{} may focus on a subset of modes of \pdata{}, ignoring other modes that may have significant mass but are not taken into account by the expectation over \pmain{}.

\subsection{MLE methods} \label{ap:mle_algorithm}
When optimizing \eqref{P:mle} on few data samples, the model may overfit the data.
One method to avoid overfitting is to add random noise to the samples~\cite{bishopTrainingNoiseEquivalent1995}.
We present a generalized version of \eqref{P:mle} to include this as
\begin{prob}
    \begin{aligned}\minimize_{\parammain\in\R^{k}} & -\frac{1}{N}\sum_{i=1}^{N}\log\pmain\left(x_{i}+n_{i}\right)-w_{H}H\left(\pmain\right)\\
 & \triangleq-\frac{1}{N}\sum_{i=1}^{N}\log\pmain\left(x_{i}+n_{i}\right)+\frac{1}{N}w_{H}\sum_{i=1}^{N}\log\pmain\left(y_{i}\right)\text{.}
\end{aligned}
\end{prob}
Here, \(x_{i}\sim\pdata\) are data samples, \(n_{i}\sim\mathcal{N}\) are samples from a noise distribution, e.g., Normal distribution, and \(y_{i}\) are samples from \pmain{} to estimate the entropy with the weight \(w_{H}\).
While maximizing the entropy of \pmain{} is not common in density estimation, we include it since the reverse KL term that we optimize in \eqref{P:final} also includes the entropy of \pmain{}.
By regularizing the entropy of \pmain{} we can compare the effect of the entropy term.

\subsection{Generative adversarial networks}
\label{ap:related_work_gan}
When minimizing only the empirical forward \KL{}, the model learns to give high probability to the training samples, but may assign high probability to regions where \pdata{} is low, i.e., have low precision.
This is due to the asymmetry of the KL divergence \(\KL\left(p\parallel q\right)\neq\KL\left(q\parallel p\right)\), which relies on sampling from only one of the distributions.
Minimizing a symmetric divergence may avoid this pitfall, but generally requires access to the unknown \pdata{}.
However, some symmetric divergences can be expressed as a variational representation without direct access to \pdata{}.
For example, minimizing any \fdivname{}, a divergence \fdiv{} defined by a function \(f\), can be formulated as the adversarial min-max objective~\citep{nguyenEstimatingDivergenceFunctionals2010,nowozinFGANTrainingGenerative2016}
\begin{prob}[P-FGAN]
    \min_{\parammain}\fdiv\left(\pdata\parallel\pmain\right)=\min_{\parammain}\max_{\paramproxy}\left(\E_{x\sim\pdata}\left[f\left(V_{\paramproxy}\left(x\right)\right)\right]+\E_{x\sim\pmain}\left[-f^{*}\left(V_{\paramproxy}\left(x\right)\right)\right]\right),
\end{prob}
where \(V_{\paramproxy}\) is a parameterized function and \(f^{*}\) is the convex conjugate of \(f\).
The most common applications of this formulation are the generative adversarial networks (GAN)~\citep{goodfellowGenerativeAdversarialNets2014} minimizing the Jensen-Shannon divergence \(D_{\text{JS}}\) as
\begin{prob}[P-GAN]
    \min_{\parammain}D_{\text{JS}}\left(\pdata\parallel\pmain\right)=\min_{\parammain}\max_{\paramproxy}\left(\E_{x\sim\pdata}\left[\log V_{\paramproxy}\left(x\right)\right]+\E_{x\sim\pmain}\left[\log\left(1-V_{\paramproxy}\left(x\right)\right)\right]\right)\text{,}
\end{prob}
and WGAN~\citep{arjovskyWassersteinGenerativeAdversarial2017} which minimizes the Wasserstein distance\footnote{Wasserstein distance is not an \(f\)-divergence, but can also be expressed as a variational problem with a 1-Lipschitz constraint over \(g_{\paramproxy}\).} \(D_{\text{W}}\) as
\begin{prob}[P-WGAN]
    \min_{\parammain}D_{\text{W}}\left(\pdata\parallel\pmain\right)=\min_{\parammain}\max_{\paramproxy,\left\Vert V_{\paramproxy}\right\Vert _{\text{L}}\leq1}\left(\E_{x\sim\pdata}\left[V_{\paramproxy}\left(x\right)\right]-\E_{x\sim\pmain}\left[V_{\paramproxy}\left(x\right)\right]\right)\text{,}
\end{prob}
where \(\left\Vert V_{\paramproxy}\right\Vert _{\text{L}}\leq1\) restricts \(V_{\paramproxy}\) to be 1-Lipschitz.
The advantages of adversarial training are the efficient sampling from the generator \(\pmain\), and that the discriminator \(V_{\paramproxy}\) can be used as an EBM~\citep{cheYourGANSecretly2020,ben-dovAdversarialLikelihoodEstimation2024}.
Yet, GANs' adversarial min-max game between the two models may not always converge to a stable solution, may have vanishing gradients or suffer mode collapse~\citep{arjovskyPrincipledMethodsTraining2017,meschederWhichTrainingMethods2018,farniaGANsAlwaysHave2020}.
In this paper we propose a non-adversarial objective to train a symmetrized divergence.

\subsection{Minimizing the forward and reverse KL}
Prior attempts to explicitly minimize both the forward and reverse KL divergences include the \(\alpha\)-bridge~\citep{zhaoBridgingMaximumLikelihood2020}.
This method leverages the definition of an \(\alpha\)-divergence~\citep{cichockiFamiliesAlphaBeta2010} which converges to the forward and reverse KL in the limits \(\alpha\rightarrow0^{+}\) and \(\alpha\rightarrow1^{-}\) by training the model in three steps.
First, the algorithm minimizes the forward KL by maximizing the evidence lower bound (ELBO), similarly to variational autoencoders (VAE).
Second, the algorithm increases \(\alpha\) according to a fixed schedule and minimizes the corresponding \(\alpha\)-divergence.
Third, the algorithm minimizes the reverse KL through adversarial training (as the \(f\)-divergence in \cref{ap:related_work_gan}).
Our method (\cref{sec:algorithm}) avoids the ELBO approximation and adversarial training by jointly optimizing both divergences in a single objective.
Additionally, rather than a fixed schedule to balance between the two divergences, our method dynamically adjusts the weight of each direction during optimization.

\subsection{Jointly training EBM and NF}
\label{ap:related_work_joint}
NFs are fast samplers but have low expressiveness compared to EBMs, which are flexible, but require a slow MCMC sampling process.
This contrast encouraged research into jointly training these models so each model's strength covers for the other's weakness.
\citet{xieTaleTwoFlows2022} propose to fit an NF to an EBM, which will then generate an initial starting point for the Langevin dynamics in the training of the EBM.
This approach accelerates the training of the EBM but keeps the MCMC process, albeit shortening the mixing time.
Other approaches remove the MCMC by installing the EBM and NF in a GAN setting~\citep{groverFlowGANCombiningMaximum2018,gaoFlowContrastiveEstimation2020,ben-dovAdversarialLikelihoodEstimation2024}.
While these works avoid the slow EBM sampling, they inherit the instability of adversarial training.

\section{Algorithm}
\label{ap:algorithm}

\subsection{NF as proxy model}
\label{ap:nf_algorithm}
Here we assume that both the main \pmain{} and proxy \pproxy{} models are NFs, and can automatically compute the gradient of their log-likelihoods with respect to their parameters.
We derive the gradients to the following version of \eqref{P:empirical}, where for convenience we write the empirical means as expectations,
\begin{prob}
   \begin{aligned}\minimize_{\substack{\parammain,\paramproxy\\
\alleps,\alldeltas\geq0
}
} &  &  & \ufor^{2}-\vfor+\urev^{2}-\vrev+\uprx^{2}-\vprx\\
\subjectto &  &  & \E_{x\sim\pdata}\left[-\log\pmain\left(x\right)\right]\leq\ufor-\vfor^{2}\\
 &  &  & \E_{y\sim\pmain}\left[\log\pmain\left(y\right)-\log\pproxy\left(y\right)\right]\leq\urev-\vrev^{2}\\
 &  &  & \E_{x\sim\pdata}\left[-\log\pproxy\left(x\right)\right]\leq\uprx-\vprx^{2}\text{.}
\end{aligned}
\end{prob}
For consistency, we denote samples from \pdata{} as \(x\) and samples from \pmain{} as \(y\).
The corresponding Lagrangian is
\begin{equation}
    \begin{aligned}\Ls\left(\parammain,\paramproxy,\alleps,\alldeltas,\lagrangevector\right) & =\ufor^{2}-\vfor+\urev^{2}-\vrev+\uprx^{2}-\vprx\\
 & \phantom{=}+\lmbfor\left(\E_{x\sim\pdata}\left[-\log\pmain\left(x\right)\right]-\ufor+\vfor^{2}\right)\\
 & \phantom{=}+\lmbrev\left(\E_{y\sim\pmain}\left[\log\pmain\left(y\right)-\log\pproxy\left(y\right)\right]-\urev+\vrev^{2}\right)\\
 & \phantom{=}+\lmbprx\left(\E_{x\sim\pdata}\left[-\log\pproxy\left(x\right)\right]-\uprx+\vprx^{2}\right)\\
 & =\ufor^{2}-\vfor+\urev^{2}-\vrev+\uprx^{2}-\vprx-\lmbfor\ufor\\
 & \phantom{=}-\lmbrev\urev-\lmbprx\uprx+\lmbfor\vfor^{2}+\lmbrev\vrev^{2}+\lmbprx\vprx^{2}\\
 & \phantom{=}+\E_{x\sim\pdata}\left[-\lmbfor\log\pmain\left(x\right)-\lmbprx\log\pproxy\left(x\right)\right]\\
 & \phantom{=}+\lmbrev\E_{y\sim\pmain}\left[\log\pmain\left(y\right)-\log\pproxy\left(y\right)\right]
\end{aligned}
\end{equation}

We optimize this problem using gradient ascent-descent using the gradients below.

\subsubsection[Closed form solution for epsilon and delta]{Closed form solution for \(\alleps,\alldeltas\)}
\label{ap:closed_form_epsilon}
Given \lagrangevector{}, equating the derivatives \(\nabla_{\alleps}\Ls,\nabla_{\alldeltas}\Ls\) to zero, results in a closed form solution, as
\begin{equation}
    \label{eq:closed_form_u}
    \begin{aligned}0 & =\nabla_{\alleps}\Ls & \qquad &  & 0 & =\nabla_{\alldeltas}\Ls\\
 & =\nabla_{\alleps}\left[\alleps^{2}-\lagrangevector\alleps\right] & \qquad &  &  & =\nabla_{\alldeltas}\left[\lagrangevector\alldeltas^{2}-\alldeltas\right]\\
 & =2\alleps-\lagrangevector & \qquad &  &  & =2\lagrangevector\alldeltas-1\\
\alleps & =\frac{1}{2}\lagrangevector & \qquad &  & \alldeltas & =\frac{1}{2\lagrangevector}
\end{aligned}
\end{equation}

\subsubsection[Gradient for theta]{Gradient for \parammain{}}
\label{ap:gradient_main}
We first derive helper gradients.
The first helper gradient is 
\begin{equation}
    \begin{aligned}\nabla_{\parammain}\E_{y\sim\pmain}\left[\log\pproxy\left(y\right)\right] & =\nabla_{\parammain}\int\pmain\left(y\right)\log\pproxy\left(y\right)\text{d}y\\
 & =\int\nabla_{\parammain}\pmain\left(y\right)\log\pproxy\left(y\right)\text{d}y\\
 & =\int\pmain\left(y\right)\log\pproxy\left(y\right)\nabla_{\parammain}\log\pmain\left(y\right)\text{d}y\\
 & =\E_{y\sim\pmain}\left[\log\pproxy\left(y\right)\nabla_{\parammain}\log\pmain\left(y\right)\right]
\end{aligned}
\end{equation}
Here we use the identity \(\nabla_{\parammain}\pmain\left(y\right)=\pmain\left(y\right)\nabla_{\parammain}\log\pmain\left(y\right)\).
Then,
\begin{equation}
    \begin{aligned}\nabla_{\parammain}\E_{y\sim\pmain}\left[\log\pmain\left(y\right)\right] & =\nabla_{\parammain}\int\pmain\left(y\right)\log\pmain\left(y\right)\text{d}y\\
 & =\int\nabla_{\parammain}\pmain\left(y\right)\log\pmain\left(y\right)\text{d}y+\int\pmain\left(y\right)\nabla_{\parammain}\log\pmain\left(y\right)\text{d}y\\
 & =\int\pmain\left(y\right)\log\pmain\left(y\right)\nabla_{\parammain}\log\pmain\left(y\right)\text{d}y+\cancel{\int\nabla_{\parammain}\pmain\left(y\right)\text{d}y}\\
 & =\E_{y\sim\pmain}\left[\log\pmain\left(y\right)\nabla_{\parammain}\log\pmain\left(y\right)\right]
\end{aligned}
\end{equation}
Where the canceled term is due to \(\int\nabla_{\parammain}\pmain\left(y\right)\text{d}y=\nabla_{\parammain}\int\pmain\left(y\right)\text{d}y=\nabla_{\parammain}1=0\).

Finally, we can plug these in and get the gradient
\begin{equation}
    \label{eq:full_gradnf}
    \begin{aligned}\nabla_{\parammain}\Ls & =\nabla_{\parammain}\left[\E_{x\sim\pdata}\left[-\lmbfor\log\pmain\left(x\right)\right]+\lmbrev\E_{y\sim\pmain}\left[\log\pmain\left(y\right)-\log\pproxy\left(y\right)\right]\right]\\
 & =\E_{x\sim\pdata}\left[-\lmbfor\nabla_{\parammain}\log\pmain\left(x\right)\right]+\lmbrev\left(\nabla_{\parammain}\E_{y\sim\pmain}\left[\log\pmain\left(y\right)\right]-\nabla_{\parammain}\E_{y\sim\pmain}\left[\log\pproxy\left(y\right)\right]\right)\\
 & =\E_{x\sim\pdata}\left[-\lmbfor\nabla_{\parammain}\log\pmain\left(x\right)\right]+\lmbrev\E_{y\sim\pmain}\left[\left(\log\pmain\left(y\right)-\log\pproxy\left(y\right)\right)\nabla_{\parammain}\log\pmain\left(y\right)\right]\\
 & \approx\frac{1}{N}\sum_{i=1}^{N}\left[-\lmbfor\nabla_{\parammain}\log\pmain\left(x_{i}\right)\right]+\frac{1}{N}\sum_{i=1}^{N}\left[\lmbrev\left(\log\pmain\left(y_{i}\right)-\log\pproxy\left(y_{i}\right)\right)\nabla_{\parammain}\log\pmain\left(y_{i}\right)\right]
\end{aligned}
\end{equation}

\subsubsection[Gradient for psi]{Gradient for \paramproxy{}}
Here the gradient is more straightforward, as 
\begin{equation}
    \begin{aligned}\nabla_{\paramproxy}\Ls & =\nabla_{\paramproxy}\left[\E_{x\sim\pdata}\left[-\lmbprx\log\pproxy\left(x\right)\right]+\lmbrev\E_{y\sim\pmain}\left[-\log\pproxy\left(y\right)\right]\right]\\
 & =\E_{x\sim\pdata}\left[-\lmbprx\nabla_{\paramproxy}\log\pproxy\left(x\right)\right]+\E_{y\sim\pmain}\left[-\lmbrev\nabla_{\paramproxy}\log\pproxy\left(y\right)\right]\\
 & \approx\frac{1}{N}\sum_{i=1}^{N}\left[-\lmbprx\nabla_{\paramproxy}\log\pproxy\left(x_{i}\right)\right]+\frac{1}{N}\sum_{i=1}^{N}\left[-\lmbrev\nabla_{\paramproxy}\log\pproxy\left(y_{i}\right)\right]
\end{aligned}
\end{equation}

\subsubsection{Lagrange multipliers gradients}
The gradients of the Lagrange multipliers \lagrangevector{} are relatively straightforward:
\begin{equation}
    \begin{aligned}\nabla_{\lmbfor}\Ls & =\frac{1}{N}\sum_{i=1}^{N}\left[-\log\pmain\left(x_{i}\right)\right]-\ufor+\vfor^{2}\\
\nabla_{\lmbrev}\Ls & =\frac{1}{N}\sum_{i=1}^{N}\left[\log\pmain\left(y_{i}\right)-\log\pproxy\left(y_{i}\right)\right]-\urev+\vrev^{2}\\
\nabla_{\lmbprx}\Ls & =\frac{1}{N}\sum_{i=1}^{N}\left[-\log\pproxy\left(x_{i}\right)\right]-\uprx+\vprx^{2}
\end{aligned}
\end{equation}

\subsection{EBM as proxy model}
\label{ap:ebm_algorithm}
In practice, we formulate \cref{P:empirical} in our experiments by adding a constraint over the normalization of the EBM as
\begin{prob}\label[prob]{P:practical}
\begin{aligned}\minimize_{\substack{\parammain,\paramproxy\\
\alleps,\alldeltas\geq0
}
} &  &  & \ufor^{2}-\vfor+\urev^{2}-\vrev+\uprx^{2}-\vprx\\
\subjectto &  &  & \E_{x\sim\pdata}\left[-\log\pmain\left(x\right)\right]\leq\ufor-\vfor^{2}\\
 &  &  & \E_{y\sim\pmain}\left[\log\pmain\left(y\right)-\funcproxy\left(y\right)+\log\partition\right]\leq\urev-\vrev^{2}\\
 &  &  & \E_{x\sim\pdata}\left[-\funcproxy\left(x\right)+\log\partition\right]\leq\uprx-\vprx^{2}\\
 &  &  & 1\leq\partition\leq1+\upartition
\end{aligned}
\end{prob}
The corresponding Lagrangian is
\begin{equation}
 \label{eq:kl_lagrangian}
\begin{aligned}\Ls\left(\parammain,\paramproxy,\alleps,\alldeltas,\lagrangevector\right) & =\ufor^{2}-\vfor+\urev^{2}-\vrev+\uprx^{2}-\vprx+\lmbfor\left(\E_{x\sim\pdata}\left[-\log\pmain\left(x\right)\right]-\ufor+\vfor^{2}\right)\\
 & \phantom{=}+\lmbrev\left(\E_{y\sim\pmain}\left[\log\pmain\left(y\right)-\funcproxy\left(y\right)+\log\partition\right]-\urev+\vrev^{2}\right)\\
 & \phantom{=}+\lmbprx\left(\E_{x\sim\pdata}\left[-\funcproxy\left(x\right)+\log\partition\right]-\uprx+\vprx^{2}\right)\\
 & \phantom{=}+\lmblower\left(1-\partition\right)+\lmbupper\left(\partition-1-\upartition\right)\\
 & =\ufor^{2}-\vfor+\urev^{2}-\vrev+\uprx^{2}-\vprx+\left(\lmbrev+\lmbprx\right)\log\partition+\left(\lmbupper-\lmblower\right)\partition\\
 & \phantom{=}+\E_{x\sim\pdata}\left[-\lmbfor\log\pmain\left(x\right)-\lmbprx\funcproxy\left(x\right)\right]+\lmbrev\E_{y\sim\pmain}\left[\log\pmain\left(y\right)-\funcproxy\left(y\right)\right]\\
 & \phantom{=}-\lmbfor\ufor-\lmbrev\urev-\lmbprx\uprx+\lmblower-\lmbupper-\lmbupper\upartition+\lmbfor\vfor^{2}+\lmbrev\vrev^{2}+\lmbprx\vprx^{2}
\end{aligned}
\end{equation}
The partition function \partition{} estimation from \cref{eq:partition_is} is empirically estimated as 
\begin{equation}
    \label{eq:partition_empirical}
    \begin{aligned} & \partition\approx\frac{1}{M}\sum_{i=1}^{M}\left[e^{\funcproxy\left(y_{i}\right)-\log\pmain\left(y_{i}\right)}\right], & \text{with }y_{i}\sim\pmain.\end{aligned}
\end{equation}

We optimize this problem in \cref{alg:dual_optimization}.

\subsubsection[Closed form solution for epsilon]{Closed form solution for \(\epsvector\)}
The gradients here are the same as in \cref{ap:closed_form_epsilon}.

\subsubsection[Gradient for theta]{Gradient for \parammain{}}
The gradient here is derived the same way as in \cref{ap:gradient_main}.

\subsubsection[Gradient for psi]{Gradient for \paramproxy{}}
First we derive the gradient of \partition{}
\begin{equation}
\begin{aligned}\nabla_{\paramproxy}\partition & =\int\frac{\pmain\left(y\right)}{\pmain\left(y\right)}\nabla_{\paramproxy}e^{\funcproxy\left(y\right)}\text{d}y\\
 & =\int\pmain\left(y\right)e^{\funcproxy\left(y\right)-\log\pmain\left(y\right)}\nabla_{\paramproxy}\funcproxy\left(y\right)\text{d}y\\
 & =\E_{y\sim\pmain}\left[e^{\funcproxy\left(y\right)-\log\pmain\left(y\right)}\nabla_{\paramproxy}\funcproxy\left(y\right)\right].
\end{aligned}
\end{equation}
We use this for deriving the gradient of the Lagrangian
\begin{equation}
    \label{eq:full_gradebm}
\begin{aligned}\nabla_{\paramproxy}L & =\nabla_{\paramproxy}\left[\E_{x\sim\pdata}\left[-\lmbprx\funcproxy\left(x\right)\right]+\lmbrev\E_{y\sim\pmain}\left[-\funcproxy\left(y\right)\right]+\left(\lmbrev+\lmbprx\right)\log\partition+\left(\lmbupper-\lmblower\right)\partition\right]\\
 & =\E_{x\sim\pdata}\left[-\lmbprx\nabla_{\paramproxy}\funcproxy\left(x\right)\right]+\lmbrev\E_{y\sim\pmain}\left[-\nabla_{\paramproxy}\funcproxy\left(y\right)\right]+\left(\frac{\lmbrev+\lmbprx}{\partition}+\lmbupper-\lmblower\right)\nabla_{\paramproxy}\partition\\
 & =\E_{x\sim\pdata}\left[-\lmbprx\nabla_{\paramproxy}\funcproxy\left(x\right)\right]+\E_{y\sim\pmain}\left[-\lmbrev\nabla_{\paramproxy}\funcproxy\left(y\right)\right]\\
 & \phantom{=}+\left(\frac{\lmbrev+\lmbprx}{\partition}+\lmbupper-\lmblower\right)\E_{y\sim\pmain}\left[e^{\funcproxy\left(y\right)-\log\pmain\left(y\right)}\nabla_{\paramproxy}\funcproxy\left(y\right)\right]\\
 & =\E_{x\sim\pdata}\left[-\lmbprx\nabla_{\paramproxy}\funcproxy\left(x\right)\right]+\E_{y\sim\pmain}\left[\left(\left(\frac{\lmbrev+\lmbprx}{\partition}+\lmbupper-\lmblower\right)e^{\funcproxy\left(y\right)-\log\pmain\left(y\right)}-\lmbrev\right)\nabla_{\paramproxy}\funcproxy\left(y\right)\right]\\
 & =\frac{1}{N}\sum_{i=1}^{N}\left[-\lmbprx\nabla_{\paramproxy}\funcproxy\left(x_{i}\right)\right]+\frac{1}{N}\sum_{i=1}^{N}\left[\left(\left(\frac{\lmbrev+\lmbprx}{\partition}+\lmbupper-\lmblower\right)e^{\funcproxy\left(y_{i}\right)-\log\pmain\left(y_{i}\right)}-\lmbrev\right)\nabla_{\paramproxy}\funcproxy\left(y_{i}\right)\right]
\end{aligned}
\end{equation}
where \partition{} is estimated as in \cref{eq:partition_empirical}.

\subsubsection{Lagrange multipliers gradients}
The gradients of the Lagrange multipliers \lagrangevector{} are relatively straightforward:
\begin{equation}
    \label{eq:gradlagrange}
\begin{aligned}\nabla_{\lmbfor}\Ls & =\frac{1}{N}\sum_{i=1}^{N}\left[-\log\pmain\left(x_{i}\right)\right]-\ufor+\vfor^{2}\\
\nabla_{\lmbrev}\Ls & =\frac{1}{N}\sum_{i=1}^{N}\left[\log\pmain\left(y_{i}\right)-\funcproxy\left(y_{i}\right)+\log\partition\right]-\urev+\vrev^{2}\\
\nabla_{\lmbprx}\Ls & =\frac{1}{N}\sum_{i=1}^{N}\left[-\funcproxy\left(x_{i}\right)+\log\partition\right]-\uprx+\vprx^{2}\\
\nabla_{\lmbupper}\Ls & =\frac{1}{M}\sum_{i=1}^{M}\left[e^{\funcproxy\left(y_{i}\right)-\log\pmain\left(y_{i}\right)}\right]-1-\upartition\\
\nabla_{\lmblower}\Ls & =1-\frac{1}{M}\sum_{i=1}^{M}\left[e^{\funcproxy\left(y_{i}\right)-\log\pmain\left(y_{i}\right)}\right]
\end{aligned}
\end{equation}

\begin{algorithm}
\caption{Estimation of \(\log\partition\) according to \cref{eq:partition_is} \label{alg:logzeta_estimation}}
\begin{algorithmic}
\REQUIRE An EBM \funcproxy{}, an NF \pmain{}, number of samples \(m\)
    \STATE Sample \(\left\{ y_{i}\right\} _{1}^{m}\) from \pmain{}
    \STATE Compute the log probabilities \(\left\{ \log\pmain\left(y_{i}\right)\right\} _{1}^{m}\)
    \STATE Compute the EBM outputs \(\left\{ \funcproxy\left(y_{i}\right)\right\} _{1}^{m}\)
    \STATE \(\ensuremath{\log\partition}\leftarrow\text{LogSumExp}\left(\left\{ \funcproxy\left(y_{i}\right)-\log\pmain\left(y_{i}\right)\right\} _{1}^{m}\right)-\log m\)
    \STATE Return \(\log\partition\)
\end{algorithmic}
\end{algorithm}

\begin{algorithm}
\caption{Optimization algorithm for \cref{D:dualproblem}}
\label{alg:dual_optimization}
\begin{algorithmic}
\REQUIRE Datasets $\mathcal{D}_{\text{data}}$, learning rates $\lr$, batch size $B$, number of epochs $E$
\FOR{$e = 1$ to $E$}
    \FOR{each minibatch $\mathcal{B}_{\text{data}} \subseteq \mathcal{D}_{\text{data}}$}
        \STATE \textbf{StopGradient:} Sample \(\left|\mathcal{B}_{\text{data}}\right|\) samples \(y\) from \pmain{}
        \STATE \textbf{StopGradient:} Estimate \(\log\partition\) from \cref{alg:logzeta_estimation}
        \STATE
        \STATE Compute the log probabilities \(\log\pmain\left(y\right)\)
        \STATE Compute the output for the data samples \(\funcproxy\left(\mathcal{B}_{\text{data}}\right)\) and \pmain{} samples \(\funcproxy\left(y\right)\)
        \STATE Compute the log probability of the data under \pmain{}, i.e., \(\log\pmain\left(\mathcal{B}_{\text{data}}\right)\)
        \STATE

        \STATE Solve \(\alleps,\alldeltas\) according to \cref{eq:closed_form_u}
        \STATE Compute \(\nabla_{\paramproxy}\Ls\) from \cref{eq:full_gradebm}
        \STATE Compute \(\nabla_{\parammain}\Ls\) from \cref{eq:full_gradnf}
        \STATE Compute \(\nabla_{\lagrangevector}\Ls\)  from \cref{eq:gradlagrange}
        \STATE Run optimizer step and update \paramproxy{}, \parammain{} with gradient descent, and \lagrangevector{} with gradient ascent
        \STATE \textbf{StopGradient:} Clamp \lagrangevector{} to be non-negative \(\lagrangevector\leftarrow\max\left\{ \boldsymbol{0},\lagrangevector\right\} \)
    \ENDFOR
\ENDFOR
\STATE Return $\paramproxy^*$, $\parammain^*$
\end{algorithmic}
\end{algorithm}

\subsection{Conditional formulation}
\label{ap:conditional}
We use the following formulation for conditional probabilities
\begin{prob}
    \label[prob]{P:conditional_empirical}
    \begin{aligned}\minimize_{\paramproxy,\parammain,\boldsymbol{\epsilon}} &  &  & \ufor^{2}-\vfor+\urev^{2}-\vrev+\uprx^{2}-\vprx\\
\subjectto &  &  & -\frac{1}{N}\sum_{i=1}^{N}\log\pmain\left(\sbiparam_{i}|\sbioutcome_{i}\right)\leq\ufor-\vfor^{2}\\
 &  &  & \phantom{-}\frac{1}{N}\sum_{i=1}^{N}\KL\left(\pmain\left(\cdot|\sbioutcome_{i}\right)\parallel\pproxy\left(\cdot|\sbioutcome_{i}\right)\right)\leq\urev-\vrev^{2}\\
 &  &  & -\frac{1}{N}\sum_{i=1}^{N}\left[\funcproxy\left(\sbiparam_{i}|\sbioutcome_{i}\right)-\log\partition^{\left(\sbioutcome_{i}\right)}\right]\leq\uprx-\vprx^{2}\\
 &  &  & 1\leq\frac{1}{N}\sum_{i=1}^{N}\partition^{\left(\sbioutcome_{i}\right)}\leq1+\epsilon_{\zeta},
\end{aligned}
\end{prob}
where \(\sbiparam_{i}\) are generated by \pmain{}.
Since each condition has a different partition function, we choose to take the empirical mean of partitions over the given conditions.

\subsection{Training in high dimensions}
\label{ap:high_dim}
In high dimensions, the initial values of \pmain{} and \funcproxy{} can be in different orders of magnitude.
The initial outputs of the MLP \funcproxy{} are usually small around zero while the log probabilities of \pmain{} are larger negative values the higher the dimension.
These differences can lead to exploding gradients due to the exponent term in \cref{eq:full_gradebm}.
To mitigate this difference in order of magnitude, we add two mechanisms.
First, we add a temperature \(T\) to the EBM, which makes it easier for the EBM to output high negative values
\begin{equation}
    \pproxy\left(x\right)=\frac{1}{\partition}e^{\frac{1}{T}\funcproxy\left(x\right)}.
\end{equation}
Then, before starting training, we warm start \funcproxy{} to produce values similar to the log probabilities of \pmain{} in an iterative process.
In each iteration, we sample from \pmain{} with the log probabilities, compute the \funcproxy{} values on the samples and minimize the mean squared error between the two sets of corresponding values.
With these two additions, training starts in a stable manner with less risk of exploding gradients.

We also replace the KL term between \pmain{} and \pproxy{} with a weighted KL using the same temperature \(T\) as
\begin{equation}
    \begin{aligned}T\cdot\KL\left(\pmain\parallel\pproxy\right) & =T\E_{x\sim\pmain}\left[\log\pmain\left(x\right)-\frac{1}{T}\funcproxy\left(x\right)+\log\partition\right]\\
 & =\E_{x\sim\pmain}\left[T\log\pmain\left(x\right)-\funcproxy\left(x\right)+T\log\partition\right],
\end{aligned}
\end{equation}
which avoids the large values of \pmain{}.

\label{ap:temperature}

\section{Implementation Details}
\label{ap:technic_details}

\subsection{Model architectures}
\label{ap:architectures}
For NF and WGAN's generators, we use a neural spline flow (NSF)~\citep{durkanNeuralSplineFlows2019}, implemented with the Zuko library\footnote{https://zuko.readthedocs.io/stable/}.
For the EBMs and WGAN's discriminator we used a residual MLP comprising several residual blocks as described in \cref{alg:residual_mlp}.
Each residual block consists of several simple layers, as detailed in \cref{alg:residual_block}.
For the conditional case, we use the same architecture as the residual MLP, but replace the block with residual FiLM blocks~\citep{perezFiLMVisualReasoning2018}, as detailed in \cref{alg:cond_residual_block}.

\begin{algorithm}
\caption{Residual MLP}
\label{alg:residual_mlp}
\begin{algorithmic}
\REQUIRE Input dimension \(d\), hidden dimension \(h\), number of blocks \(n\)
\REQUIRE Input vector \(x\in\R^{d}\)
    \STATE \(x^{\prime}\leftarrow\text{Linear}\left(x\right)\)
    \COMMENT{Linear layer with output dimension \(h\) with spectral normalization}
    \STATE \(x^{\prime}\leftarrow\text{SiLU}\left(x^{\prime}\right)\)
    \COMMENT{Sigmoid Linear Unit activation}
    \FOR{\(i=1\) to \(n\)}
        \STATE \(x^{\prime}\leftarrow\text{ResidualBlock}\left(x^{\prime}\right)\)
        \COMMENT{Residual block as in \cref{alg:residual_block} with dimension \(h\)}
    \ENDFOR
    \STATE \(x^{\prime}\leftarrow\text{Linear}\left(x^{\prime}\right)\)
    \COMMENT{Linear layer with input dimension \(h\) and output dimension \(1\)}
    \STATE Return \(x^{\prime}\)
\end{algorithmic}
\end{algorithm}

\begin{algorithm}
\caption{Residual block}
\label{alg:residual_block}
\begin{algorithmic}
\REQUIRE Dimension \(d\)
\REQUIRE Input vector \(x\in\R^{d}\)
    \STATE \(x^{\prime}\leftarrow\text{Layer Normalization}\left(x\right)\)
    \STATE \(x^{\prime}\leftarrow\text{SiLU}\left(x^{\prime}\right)\)
    \COMMENT{Sigmoid Linear Unit activation}
    \STATE \(x^{\prime}\leftarrow\text{Linear}\left(x^{\prime}\right)\)
    \COMMENT{Linear layer with input and output dimension \(d\) and spectral norm}
    \STATE \(x^{\prime}\leftarrow\text{Layer Normalization}\left(x^{\prime}\right)\)
    \STATE \(x^{\prime}\leftarrow\text{SiLU}\left(x^{\prime}\right)\)
    \COMMENT{Sigmoid Linear Unit activation}
    \STATE \(x^{\prime}\leftarrow\text{Linear}\left(x^{\prime}\right)\)
    \COMMENT{Linear layer with input and output dimension \(d\) and spectral norm}
    \STATE \(x^{\prime}\leftarrow \alpha x+x^{\prime}\)
    \COMMENT{Residual connection with trainable parameter \(\alpha\)}
    \STATE \(x^{\prime}\leftarrow\text{SiLU}\left(x^{\prime}\right)\)
    \STATE Return \(x^{\prime}\)
\end{algorithmic}
\end{algorithm}

\begin{algorithm}
\caption{Conditional residual block}
\label{alg:cond_residual_block}
\begin{algorithmic}
\REQUIRE Dimension \(d\), number of FiLM layers \(m\)
\REQUIRE Input vector \(x\in\R^{d}\), condition vector \(c\in\R^{k}\)
    \STATE \(\gamma\leftarrow\text{MLP}\left(c\right)\)
    \STATE
    \COMMENT{MLP with input dimension \(k\) and output dimension \(d\), with \(m\) layers, SiLU activation}
    \STATE
    \STATE \(\beta\leftarrow\text{MLP}\left(c\right)\)
    \STATE
    \COMMENT{MLP with input dimension \(k\) and output dimension \(d\), with \(m\) layers, SiLU activation}
    \STATE
    \STATE \(x^{\prime}\leftarrow\text{MLP}\left(x\right)\)
    \STATE
    \COMMENT{MLP with input dimension \(d\) and output dimension \(d\), with \(2\) layers, SiLU activation}
    \STATE
    \STATE \(x^{\prime}\leftarrow\gamma\odot x^{\prime}+\beta\)
    \COMMENT{\(\odot\) represents element-wise multiplication}
    \STATE \(x^{\prime}\leftarrow\text{SiLU}\left(\alpha x+x^{\prime}\right)\)
    \COMMENT{Residual connection with trainable parameter \(\alpha\)}
    \STATE Return \(x^{\prime}\)
\end{algorithmic}
\end{algorithm}

\begin{algorithm}
\caption{Encoder}
\label{alg:encoder}
\begin{algorithmic}
\REQUIRE Latent dimension \(d\)
\REQUIRE Input vector \(x\in\R^{D}\)
    \STATE \(x^{\prime}\leftarrow\text{Conv2D}\left(x\right)\)
    \COMMENT{Convolution layer with 32 channels, kernel size 4, stride 2 and padding 1}
    \STATE \(x^{\prime}\leftarrow\text{ReLU}\left(x^{\prime}\right)\)
    \STATE \(x^{\prime}\leftarrow\text{Conv2D}\left(x^{\prime}\right)\)
    \COMMENT{Convolution layer with 64 channels, kernel size 4, stride 2 and padding 1}
    \STATE \(x^{\prime}\leftarrow\text{ReLU}\left(x^{\prime}\right)\)
    \STATE \(x^{\prime}\leftarrow\text{Conv2D}\left(x^{\prime}\right)\)
    \COMMENT{Convolution layer with 128 channels, kernel size 4, stride 2 and padding 1}
    \STATE \(x^{\prime}\leftarrow\text{ReLU}\left(x^{\prime}\right)\)
    \STATE \(x^{\prime}\leftarrow\text{Conv2D}\left(x^{\prime}\right)\)
    \COMMENT{Convolution layer with 256 channels, kernel size 4, stride 2 and padding 1}
    \STATE \(x^{\prime}\leftarrow\text{ReLU}\left(x^{\prime}\right)\)
    \STATE \(x^{\prime}\leftarrow\text{MLP}\left(x^{\prime}\right)\)
    \COMMENT{MLP with output dimension \(d\)}
    \STATE Return \(x^{\prime}\)
\end{algorithmic}
\end{algorithm}

\begin{algorithm}
\caption{Decoder}
\label{alg:decoder}
\begin{algorithmic}
\REQUIRE Input vector \(x\in\R^{d}\)
    \STATE \(x^{\prime}\leftarrow\text{MLP}\left(x\right)\)
    \COMMENT{MLP with output dimension \(256\times 4\times 4\)}
    \STATE \(x^{\prime}\leftarrow\text{ConvT2D}\left(x^{\prime}\right)\)
    \COMMENT{Transposed convolution: 64 channels, kernel size 4, stride 2, padding 1}
    \STATE \(x^{\prime}\leftarrow\text{ReLU}\left(x^{\prime}\right)\)
    \STATE \(x^{\prime}\leftarrow\text{ConvT2D}\left(x^{\prime}\right)\)
    \COMMENT{Transposed convolution: 32 channels, kernel size 4, stride 2, padding 1}
    \STATE \(x^{\prime}\leftarrow\text{ReLU}\left(x^{\prime}\right)\)
    \STATE \(x^{\prime}\leftarrow\text{ConvT2D}\left(x^{\prime}\right)\)
    \COMMENT{Transposed convolution: 3 channels, kernel size 4, stride 2, padding 1}
    \STATE Return \(x^{\prime}\)
\end{algorithmic}
\end{algorithm}

\subsection{Autoencoder}
\label{ap:autoencoder}
To reduce the dimensionality of CelebA, we use an autoencoder with a 100D latent space.
We use the 64x64 version of the dataset, and scale the pixel values to be between -1 and 1.
We used \cref{alg:encoder} for the encoder and \cref{alg:decoder} for the decoder, and trained for 20 epochs with a batch size of 128 and a learning rate of \(10^{-3}\) using Adam.
The objective for the autoencoder was the mean square error (MSE) loss between the input and reconstructed output.

\subsection{Optimization and evaluation}
All our code is written in Python using PyTorch\footnote{https://pytorch.org/} and Lightning\footnote{https://lightning.ai/docs/pytorch/stable/}.
For optimization, we use AdamW with parameters \((\beta_{1},\beta_{2})=(0.0, 0.9)\) and change the learning rate and batch size per experiment.

\section{Experimental details and results}
\label{ap:experimental_details}

\subsection{UCI datasets study of MLE with noise-perturbation}
\label{ap:mle_noise}

(Expanding on \cref{ssec:results_uci}.)

\begin{figure}
    \centering

    \begin{subfigure}{\linewidth}
        \centering      

        \caption{Test set NLL}
        \label{fig:uci_noise_test}
        \includegraphics[width=\linewidth]{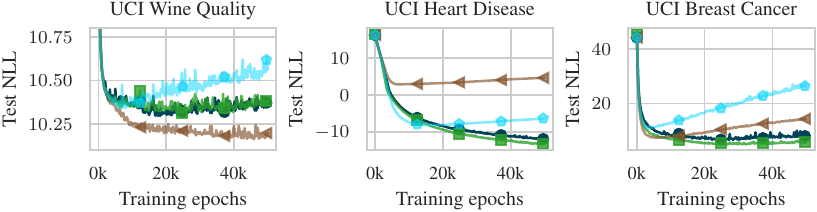}
        \includegraphics[width=0.75\linewidth]{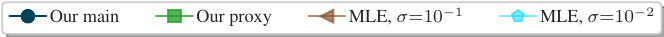}

    \end{subfigure}

    \begin{subfigure}{\linewidth}
        \centering      

        \caption{Train set NLL}
        \label{fig:uci_noise_train}
        \includegraphics[width=\linewidth]{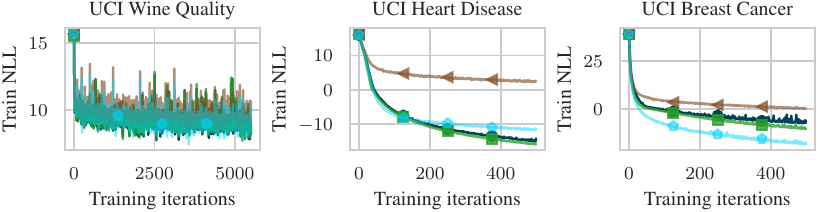}
        \includegraphics[width=0.75\linewidth]{figures/uci/test_legend.pdf}

    \end{subfigure}

    \begin{subfigure}{\linewidth}
        \centering      

        \caption{Dual variables}
        \label{fig:uci_duals}
        \includegraphics[width=\linewidth]{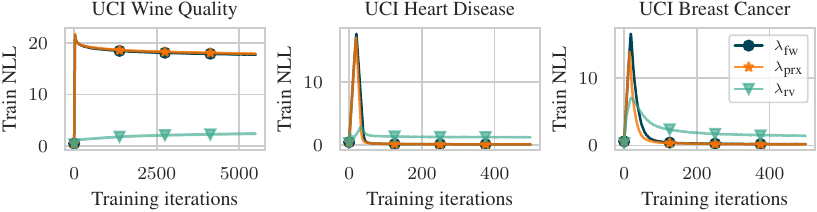}

    \end{subfigure}

    \caption{
        Result on UCI datasets. (a) NLL results on the test set. (b) NLL results on the train set. (c) Values of the dual variables during training.
    }
    
\end{figure}

We used public datasets from the machine learning repository\footnote{https://archive.ics.uci.edu/datasets} of the University of California, Irvine~(UCI).
Although these datasets are commonly used for classification, we ignore the labels and consider only density estimation over the features.
Before training, we apply the same preprocessing pipeline to all UCI datasets:
\begin{enumerate}
    \item Remove rows with NaN (Not a Number) values.
    \item Any non-numerical features, such as categorical features, are encoded as integers and perturbed with uniform random noise. The noise is in the range \(\left[-0.2,0.2\right]\), and is predetermined to be consistent across all runs.
    \item Normalize each feature to have zero mean and unit variance across all samples.
    \item Randomly split the data into a train set of 80\% of the samples and a test set of 20\% of the samples.
\end{enumerate}
We trained the models only on the train set, and used the test set only for evaluation.
The datasets we used are:
\begin{itemize}
    \item \emph{Wine Quality}~\citep{cortezUsingDataMining2009} contains 6,497 samples of 11 physicochemical properties of red and white wines.
    \item \emph{Heart Disease}~\citep{detranoInternationalApplicationNew1989} comprises 303 patients and results of 11 tests to predict coronary artery disease.
    \item \emph{Breast Cancer Wisconsin (Diagnostic)}~\citep{streetNuclearFeatureExtraction1993} contains 569 samples of 30 features for diagnosing breast cancer.
\end{itemize}

\begin{table}
    \caption{UCI datasets and training details.}
    \label{tbl:uci}
    \centering
    \begin{tabular}{@{}lcccccc@{}}
\toprule
Dataset &
  \multicolumn{1}{l}{Features} &
  \multicolumn{1}{l}{\begin{tabular}[c]{@{}l@{}}Train\\ samples\end{tabular}} &
  \multicolumn{1}{l}{\begin{tabular}[c]{@{}l@{}}Test\\ samples\end{tabular}} &
  \multicolumn{1}{l}{\begin{tabular}[c]{@{}l@{}}NSF\\ transforms\end{tabular}} &
  \multicolumn{1}{l}{\begin{tabular}[c]{@{}l@{}}NSF\\ bins\end{tabular}} &
  \multicolumn{1}{l}{\begin{tabular}[c]{@{}l@{}}Learning\\ rate\end{tabular}} \\ \midrule
Wine Quality  & 11 & 5197 & 1300 & 1 & 2 & \(10^{-4}\) \\
Heart Disease & 11 & 242  & 61   & 4 & 8 & \(10^{-5}\) \\
Breast Cancer & 30 & 455  & 114  & 2 & 8 & \(10^{-4}\) \\ \bottomrule
\end{tabular}
\end{table}

For each dataset we used a different NSF architecture, as detailed in \cref{tbl:uci}.
We used the same learning rate for the main model and the proxy, i.e., \(\lr_{\parammain}=\lr_{\paramproxy}\) and for the dual variables a learning rate of \(\lr_{\lambda} = 10^{-2}\).
All methods were trained for 50,000 epochs.

Our framework used the same architecture for the main and proxy model, without sharing parameters.
We train our models as in \cref{ap:nf_algorithm}.
We trained multiple variants of MLE (\cref{ap:mle_algorithm}):
\begin{itemize}
    \item Pure MLE: we trained \pmain{} only by minimizing the NLL of the train set.
    \item MLE with noise-perturbation: we trained \pmain{} by minimizing the NLL of the train set, where we added Gaussian noise with zero mean and standard deviations \(\sigma\in\left\{ 10^{-1},10^{-2}\right\} \).
    \item MLE with entropy regularization: we trained \pmain{} by minimizing the NLL of the train set with an entropy regularization term to maximize the entropy of \pmain{}. We computed the entropy by sampling from \pmain{} and evaluate the samples' log-probability under \pmain{}. We used a regularization coefficient of 0.1. 
\end{itemize}

In \cref{fig:uci_noise_test} we compare the test set NLL of different levels of noise-perturbation.
These results show that changing the \(\sigma\) of the noise drastically affects the results, determining whether the NF will overfit or not, and in one case, even outperforming our method.
We find evidence of overfitting in \cref{fig:uci_noise_train}, which compares the train set NLL, and shows that the train NLL of all methods keep improving, although some of their test NLL gets worse.

Finally, in \cref{fig:uci_duals} we show the evolution of the dual variables throughout training.

\subsection{Comparison on 40-component Gaussian mixture}
\label{ap:gmm40}
(Details for \cref{ssec:ebm_as_proxy})

\begin{figure}
    \centering

    \begin{subfigure}{\linewidth}
        \centering      

        \caption{Jeffreys divergence}
        \includegraphics[width=0.85\linewidth]{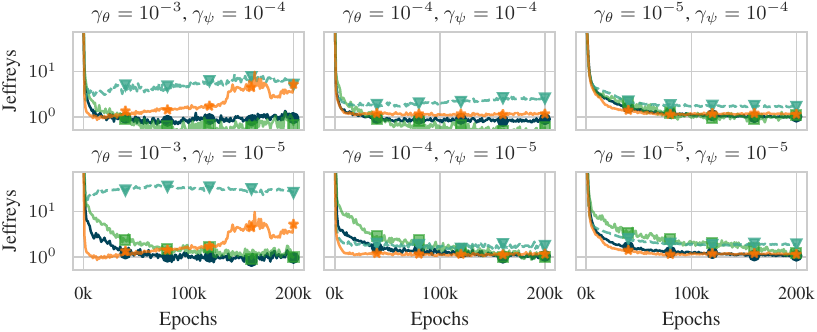}
        \includegraphics[width=0.5\linewidth]{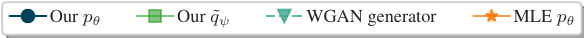}

    \end{subfigure}

    \begin{subfigure}{\linewidth}
        \centering      

        \caption{NLL}
        \includegraphics[width=0.85\linewidth]{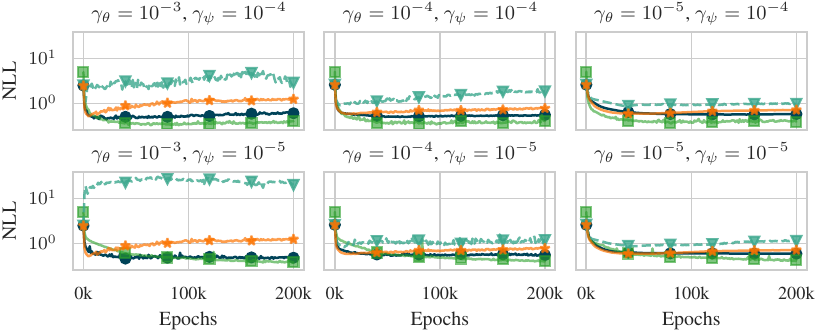}
        \includegraphics[width=0.5\linewidth]{figures/wgans/divergence_legend.pdf}

    \end{subfigure}

    \begin{subfigure}{\linewidth}
        \centering      

        \caption{Dual variables}
        \includegraphics[width=0.85\linewidth]{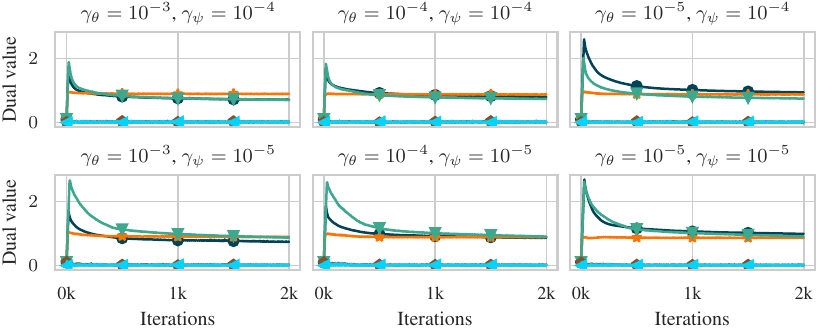}
        \includegraphics[width=0.5\linewidth]{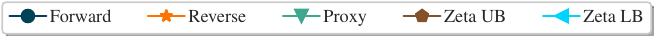}

    \end{subfigure}

    \caption{
        Result on 2D GMM with 40 components. (a) Jeffreys divergence computed on the test set. (b) NLL computed on the test set. (c) Values of the dual variables during training.
    }
    \label{fig:ap_wgan}
    
\end{figure}

In \cref{sec:algorithm} we compare our method with the weighted sum formulation, NF and WGAN.
The 2D distribution we used for the comparison, inspired by \citet{midgleyFlowAnnealedImportance2023}, consisted of a Gaussian mixture with 40 components.
The center of each component is drawn uniformly between 0 and 1, and each component has a standard deviation of 0.05.
For the train set we used 800 samples, and for the test set 10,000 samples.
We trained all models for 200,000 iterations with full batches of the whole train set.
For the Lagrange multipliers we used a learning rate of \(10^{-3}\), and set \(\upartition=0.01\).
We also set a weight decay of \(10^{-4}\) and gradient clipping of \(5.0\) for all methods to keep the GANs stable.

For the main model \pmain{} and the WGAN generator we used a neural spline flow (\cref{ap:architectures}) with 4 transforms, 6 bins and hidden channels \(\left(32,32\right)\).
For the proxy model \pproxy{} and WGAN's discriminator we used a residual MLP (\cref{alg:residual_mlp}) with 2 residual blocks and a hidden dimension of 128.

In the comparison with the weighted sum formulation (\cref{P:penalty}) we used 25 different weight configurations for \wfw{} and \wprx{}, where each weight \(\wfw,\wprx\in\left\{ 0.1,0.25,0.5,0.75,0.9\right\} \).
For both formulations, we used a learning rate (LR) of \(\lr_{\parammain}=10^{-4}\) for \pmain{} and \(\lr_{\paramproxy}=10^{-4}\) for \pproxy{}, and we estimated \partition{} with \(M=1000\) samples from \pmain{}.

When comparing with WGAN, we used several learning rates, i.e., \(\lr_{\parammain}\in\left\{ 10^{-3},10^{-4},10^{-5}\right\}\) for the main model \pmain{} and the generator, and \(\lr_{\paramproxy}\in\left\{ 10^{-4},10^{-5}\right\}\) for the proxy model \pproxy{} and the discriminator.
We trained the WGAN with 5 gradient steps per iteration for the discriminator and set the gradient penalty~\citep{gulrajaniImprovedTrainingWasserstein2017} with a coefficient of 0.1.
We present the results of all learning rates in \cref{fig:ap_wgan}.
In the main text we include only \(\lr_{\parammain}\in\left\{ 10^{-3},10^{-5}\right\}\) and \(\lr_{\paramproxy}=10^{-5}\).

\begin{figure}
    \centering

    \begin{subfigure}{0.45\linewidth}
        \centering
        \caption{Concentric rings NLL}
        \includegraphics[width=\linewidth]{figures/density/vals_concentric.pdf}

    \end{subfigure}
    \hfill
    \begin{subfigure}{0.45\linewidth}
        \centering      
        \caption{Concentric rings density}
        \includegraphics[width=\linewidth]{figures/density/concentric.png}

    \end{subfigure}

    \begin{subfigure}{0.45\linewidth}
        \centering
        \caption{Gaussian mixture grid NLL}
        \includegraphics[width=\linewidth]{figures/density/vals_grid.pdf}

    \end{subfigure}
    \hfill
    \begin{subfigure}{0.45\linewidth}
        \centering      
        \caption{Gaussian mixture grid density}
        \includegraphics[width=\linewidth]{figures/density/grid.png}

    \end{subfigure}

    \begin{subfigure}{0.45\linewidth}
        \centering
        \caption{Gaussian mixture ring NLL}
        \includegraphics[width=\linewidth]{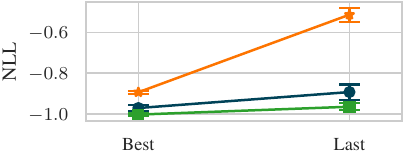}

    \end{subfigure}
    \hfill
    \begin{subfigure}{0.45\linewidth}
        \centering
        \caption{Gaussian mixture ring density}
        \includegraphics[width=\linewidth]{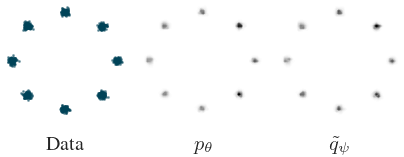}

    \end{subfigure}

    \begin{subfigure}{0.45\linewidth}
        \centering
        \caption{Spiral NLL}
        \includegraphics[width=\linewidth]{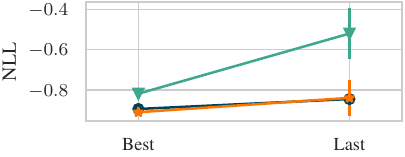}

    \end{subfigure}
    \hfill
    \begin{subfigure}{0.45\linewidth}
        \centering      
        \caption{Spiral density}
        \includegraphics[width=\linewidth]{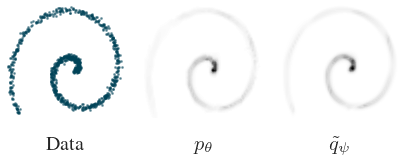}

    \end{subfigure}

    \begin{subfigure}{0.45\linewidth}
        \centering
        \caption{Moons NLL}
        \includegraphics[width=\linewidth]{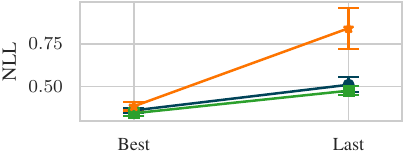}

    \end{subfigure}
    \hfill
    \begin{subfigure}{0.45\linewidth}
        \centering      
        \caption{Moons density}
        \includegraphics[width=\linewidth]{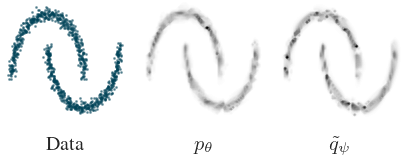}

    \end{subfigure}

    \caption{
    Our framework is able to accurately learn the density of various 2D datasets.
    The left column presents the lowest NLL each model achieved during training and the NLL at the end of training, demonstrating that our method outperforms NF in terms of values and consistency.
    The right column qualitatively displays the finite dataset (left) and the learned density values from of \pmain{} and the quasi-normalized \(\tilde{q}_{\paramproxy}\) (middle and right respectively).
    The qualitative density maps show that both models perfectly capture the shape and all modes.
    }
    \label{fig:appendix_density}
\end{figure}

\subsection{Density estimation}
\label{ap:shapes}
(Details for \cref{ssec:density})

In the density estimation experiments in \cref{ssec:density} we tested 5 different distributions, which can be seen in \cref{fig:appendix_density}.
For each dataset we used a train set of 1000 samples and a test set of 10000 samples and trained for 25000 full-batch iterations.
We used a LR of \(\lr_{\parammain}=10^{-3}\) for \pmain{} and NF, and LR of \(\lr_{\paramproxy}=10^{-4}\) for \pproxy{}.
We estimated \partition{} with \(M=1000\) samples from \pmain{}.
For the Lagrange multipliers we used a learning rate of \(10^{-4}\), while the Lagrange multipliers corresponding to the \partition{} normalization constraints were trained with a learning rate of \(10^{-3}\), where \(\upartition=0.1\).
The initial values for the KL Lagrange multipliers was set to 0.01.

To create the concentric rings dataset, we sample \(N\) points from a standard normal distribution, normalize them to have unit norm, and scale them by either \(r_{1}=0.2\) or \(r_{2}=0.6\) with 0.5 probability.
We then added uniform random noise \(\mathcal{U}(0, 0.1)\) to each point.
To create the Gaussian mixture grid, we created a Gaussian mixture distribution of a 5x5 grid of centers between -1 and 1 which we then scaled to have zero mean and unit variance, each with a standard deviation of 0.05.
To create the Gaussian mixture ring, we created a Gaussian mixture distribution of 8 components equidistant on a unit circle which we then scaled to have zero mean and unit variance. Each component has a standard deviation of 0.05.
To create the spiral distribution, we sampled \(N\) points uniformly \(t\sim\mathcal{U}\left(0,10\right)\), and created points \((x,y)\) with \(x=t\cos(t)\) and \(y=t\sin(t)\). We then added Gaussian noise with standard deviation of 0.02 to each point.
For the moons dataset we used the \textit{make\_moons} function from the scikit-learn package\footnote{https://scikit-learn.org/}.

To draw the qualitative results, we divided the visualized region into \(64\times 64\) pixels and computed the density value of each pixel as \(v=p\left(x\right)\cdot A\), with \(A\) being the area of each pixel.
We then plot the filled contours of the density values.

\subsection{Image sampling}
\label{ap:img_sampling}

\begin{figure*}[t]
    \centering
    \begin{subfigure}{0.49\linewidth}
        \centering
        \caption{Our FID: 48.07}
        \includegraphics[width=\linewidth]{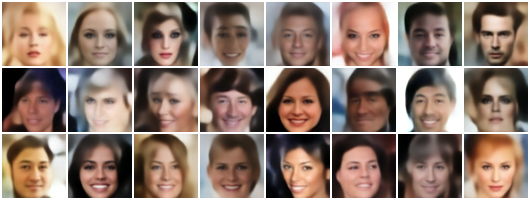}

    \end{subfigure}
    \hfill
    \begin{subfigure}{0.49\linewidth}
        \centering      
        \caption{NF FID: 48.99}
        \includegraphics[width=\linewidth]{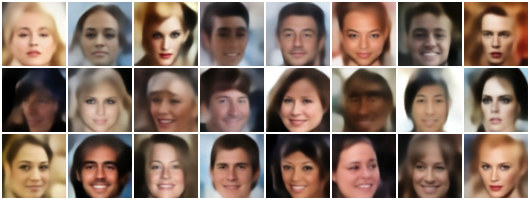}

    \end{subfigure}

    \caption{
        Our \pmain{} learns a comparable 100D encoded space of CelebA as NF.
        (a) reports the FID and qualitative samples generated from our \pmain{}, and (b) reports the same for NF.
        We used the same seed to generate the samples, which explains the similarity between the images.
    }
    \label{fig:celeba}
\end{figure*}

We illustrate the sampling quality of \pmain{} by learning to sample from the image dataset CelebA~\citep{liuDeepLearningFace2015}.
The main problem when training on a high-dimensional space is that the gradient of \paramproxy{} may explode due to the exponential in \(\nabla_{\paramproxy} \Ls\) (see \cref{ap:algorithm}).
We mitigate this issue by taking three steps.
First, we train an autoencoder on the training dataset and learn 100D encodings that serve as the training data for our models.
Second, we add a temperature parameter to the EBM that scales the density values to the correct order of magnitude.
Third, before training we warm start the EBM by training it to have a similar density as the NF.
\cref{ap:high_dim} details the implementation of these steps and below we detail the training details.
To compare between our method and a pure NF, we use the Fréchet inception distance (FID)~\citep{heuselGANsTrainedTwo2017}, a common metric to measure the generation quality of a model.
\cref{fig:celeba} displays the decoded samples from \pmain{} and an NF and their FID values.
Our model gets comparable FID to NF for mainly two reasons. First, our NF does not overfit on CelebA because the dataset contains 200k images.
Second, the estimation of \partition{} has high variance since the importance sampling estimator for \partition{} requires a large number of samples in a 100D space to be accurate, and we only use 5000 samples every 100 iterations.

We train our models on the encoded 100D space of CelebA~\citep{liuDeepLearningFace2015}.
We use an autoencoder to reduce dimensionality as described in \cref{ap:autoencoder}.
For \pmain{} and NF we used a neural spline flow (\cref{ap:architectures}) with 6 transforms and 16 bins.
For \pproxy{} we used a residual MLP (\cref{alg:residual_mlp}) with 5 residual blocks and a hidden dimension of 1024.
For training our models, we used a temperature 0.01, as described in \cref{ap:high_dim}, and used 5000 samples to estimate \partition{}.
To save time, we sampled the 5000 samples only every 100 iterations.
We trained all models for 40 epochs with batch size 128.
The LR for \pmain{} and NF was \(10^{-3}\), and for \pproxy{} it was \(10^{-6}\).
After every epoch, we evaluated the FID\footnote{We adapted the code from https://github.com/mseitzer/pytorch-fid.
} with 50,000 samples and report the best FID that \pmain{} and NF achieved during training.
For the Lagrange multipliers we used a learning rate of \(10^{-6}\), while the Lagrange multipliers corresponding to the \partition{} normalization constraints were trained with a learning rate of \(10^{-4}\), where \(\epsilon_{\zeta}=100\).

\subsection{Simulation-based inference}
\label{ap:sbi}
(Details for \cref{ssec:sbi})

For SBI, we used two benchmark datasets: two moons~\citep{greenbergAutomaticPosteriorTransformation2019}, Gaussian mixture~\citep{sissonSequentialMonteCarlo2007}.
Each benchmark defines a prior and a simulator.
\paragraph{Gaussian mixture} The parameter dimension is \(\sbiparam\in\R^{2}\) and the outcome dimension is \(\sbioutcome\in\R^{2}\).
The prior is a uniform distribution \(p_{\sbiparam}=\mathcal{U}\left(\left[0.5,1\right]^{2}\right)\) and the simulator is a Gaussian mixture of 2 components \(p\left(sbioutcome|\sbiparam\right)=0.5\mathcal{N}\left(\sbiparam,I\right)+0.5\mathcal{N}\left(\sbiparam,0.01I\right)\), where \(I\) is the identity matrix.
\paragraph{Two moons} The parameter dimension is \(\sbiparam\in\R^{2}\) and the outcome dimension is \(\sbioutcome\in\R^{2}\).
The prior is a uniform distribution \(p_{\sbiparam}=\mathcal{U}\left(\left[-1,1\right]^{2}\right)\).
The simulator is \[\sbioutcome|\sbiparam=\left(r\cos\left(\alpha\right)+0.25,r\sin\left(\alpha\right)\right)+\left(-\left|\sbiparam_{1}+\sbiparam_{2}\right|/\sqrt{2},\left(-\sbiparam_{1}+\sbiparam_{2}\right)/\sqrt{2}\right),\] where \(\alpha\sim\mathcal{U}\left(-\frac{\pi}{2},\frac{\pi}{2}\right)\) and \(r\sim\mathcal{N}\left(0.1,0.01^{2}\right)\).

To evaluate the C2ST, we generate 5000 samples from the posterior using rejection sampling.
For each dataset we defined a test condition \(\sbioutcome_{0}\) and generated samples by sampling from the prior, running the simulator and keeping the first 5000 samples that were close to \(\sbioutcome_{0}\).
For the Gaussian mixture we set \(\sbioutcome_{0}=(0.75,0.75)\) and for the two moons \(\sbioutcome_{0}=(0,0)\).
We also generate 5000 samples from the trained models by conditioning on \(\sbioutcome_{0}\).
We split the total of 10000 into a train and test set with a 0.3 split, and train a classifier\footnote{HistGradientBoostingClassifier from https://scikit-learn.org} with positive labels for the ground truth points and negative labels for the model points, and use its accuracy on the test set as the C2ST metric.

To generate from \pproxy{}, we followed the MCMC sampling in \cref{eq:mcmcsampling}.
We used 2000 steps with a step size of \(10^{-4}\). The initial points were sampled from \pmain{}.

For \pmain{} and NF we used 4 transforms and 8 bins. For \pproxy{} we used a different architecture for each dataset.
For the Gaussian mixture, we used a conditional residual MLP (\cref{alg:cond_residual_block}) with 6 residual blocks and a hidden dimension of 64 and 2 FiLM layers.
For the two moons, we used 3 residual blocks, each with 128 channels and 2 FiLM layers.
We trained with different number of simulations (data points), but used a batch size of 1000 for all experiments.
We used  LR of \(10^{-4}\) for \pmain{} and NF, and LR of \(10^{-5}\) for \pproxy{}.
The Lagrange multipliers LR was set to \(10^{-6}\), while the Lagrange multipliers corresponding to the \partition{} normalization constraints were trained with a learning rate of \(10^{-2}\), where \(\epsilon_{\zeta}=30\).
The initial values for the KL Lagrange multipliers was set to 0.01.

\section{Compute time}
\label{ap:compute_time}
\begin{figure}
    \centering
    \begin{subfigure}{0.49\linewidth}
        \centering      
        \caption{Average time per epoch}
        \label{fig:runtime_time}

        \includegraphics[width=\linewidth]{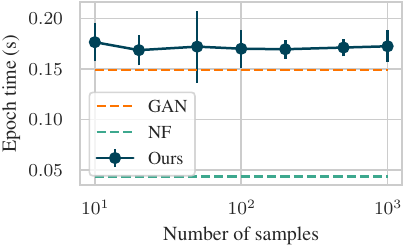}

    \end{subfigure}
    \hfill
    \begin{subfigure}{0.49\linewidth}
        \centering      
        \caption{Lowest NLL}
        \label{fig:runtime_nll}

        \includegraphics[width=\linewidth]{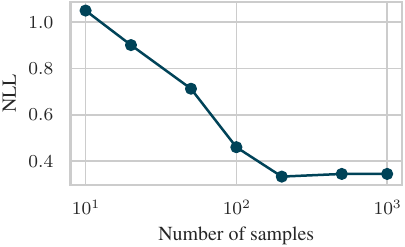}

    \end{subfigure}

    \caption{
        How the number of samples for the estimation of \partition{} with importance sampling (\cref{eq:partition_is}) affects training.
    (a) The computation time for our method is slower than NF and GANs, but does not scale much with more samples.
    (b) Using more samples improves the NLL.
    }
    \label{fig:runtime}
\end{figure}

Our method requires generating \(M\) samples to estimate the partition function \partition{} with importance sampling (\cref{eq:partition_is}).
Here we study how the number of samples \(M\) affects the training time and performance.
We used our training data from the 40 components Gaussian mixture experiment from \cref{ssec:ebm_as_proxy}.
We trained all models on an NVIDIA A100 GPU and averaged the time per epoch over 100,000 epochs.

\cref{fig:runtime_time} shows that our method is indeed slower than NF, but not much slower than GANs.
It is also interesting to observe that the compute time barely increases with more samples.
This is because the importance sampling does not require gradients, and with the stop gradient (\cref{alg:dual_optimization}), the computation is efficient.
For a much larger number of samples, or higher dimensions, we expect the time per epoch to increase.

We also examine how the number of samples affects the quality of learning.
In \cref{fig:runtime_nll} we plot the best NLL achieved during training per number of samples.
It is clear that with too few samples, the inaccuracy in the estimation harms learning.

\end{document}